%% file: main.tex
\documentclass{article}

\usepackage[preprint]{preprintstyle}

\usepackage[utf8]{inputenc} 
\usepackage[T1]{fontenc}    
\usepackage{hyperref}       
\usepackage{url}            
\usepackage{booktabs}       
\usepackage{amsfonts}       
\usepackage{nicefrac}       
\usepackage{microtype}      
\usepackage{xcolor}         
\input{preamble}
\title{FoMo: Forking Moment in Generative Trajectory as a Perceptual Distance}

\author{
  Jaihyun Lew$^1$, Mingi Jung$^2$, Minjun Park$^1$, Wooseok Song$^2$, Sungroh Yoon$^{1,2,3,}$\thanks{Corresponding Author} \\
  $^1$Interdisciplinary Program in AI, Seoul National University\\
  $^2$Department of Electrical and Computer Engineering, Seoul National University\\
  $^3$AIIS, ASRI, INMC, and ISRC, Seoul National University\\
  \texttt{\{fudojhl, 2019-16552, minjunpark, cody1129, sryoon\}@snu.ac.kr}  \\
}

\begin{document}

\maketitle

\input{00_abstract}
\input{01_intro}
\input{03_preliminary}
\input{04_method}

\input{05_experiments}
\input{06_conclusion}

\begin{ack}
This work was supported by the BK21 FOUR program of the Education and Research Program for Future ICT Pioneers,Seoul National University in 2026;
the National Research Foundation of Korea (NRF) grant funded by the Korea government (MSIT) (Nos. 2022R1A3B1077720 and 2022R1A5A7083908);
and the Institute of Information \& Communications Technology Planning \& Evaluation (IITP) grant funded by the Korea government (MSIT) (No. RS-2021-II211343, Artificial Intelligence Graduate School Program, Seoul National University).
The authors also thank Yongsung Kim for valuable feedback and support.
\end{ack}

{
    \small
    \bibliographystyle{ieeenat_fullname}
    \bibliography{main}
}

\newpage
\input{appendix}


\end{document}

%% file: preamble.tex
\setcitestyle{numbers,square,comma}
\usepackage{multirow}
\usepackage{makecell}
\usepackage{pifont}
\usepackage{amsmath}
\usepackage{graphicx}
\usepackage{booktabs}
\usepackage{subcaption}
\newcommand{\cmark}{\ding{51}}
\newcommand{\xmark}{\ding{55}}

\usepackage{dsfont}

%% file: 00_abstract.tex
\begin{abstract}
Reference-based image quality assessment (IQA) metrics aim to reflect how humans perceive the perceptual distance between a pair of images. To learn how the human visual system (HVS) operates, recent reference-based IQA metrics heavily rely on human-annotated data.
Mean opinion score (MOS)-based pointwise scoring, which assigns a scalar quality value per image, is preferable for annotation but is prohibitively expensive to collect at scale and is known to be noisy due to inconsistent human judgments.
As an alternative, two-alternative forced choice (2AFC) pairwise labels have gained popularity due to their reliability and efficiency, but they capture only relative comparisons between pairs.
In this paper, we propose a fully automated data generation pipeline that generates pointwise perceptual distance labels between image pairs without any human annotation.
Our approach exploits the generative dynamics of diffusion models as a perceptual distance proxy,
where the coarse structure of an image is generated in the early timesteps and the fine details are generated in the later timesteps.
Images that fork early in the generation process share only coarse structure and are perceptually far apart; images that fork late differ only in fine detail.
We demonstrate that the diffusion trajectory aligns well with the human visual system, and use this forking moment, FoMo, as a reference-grounded distance label to supervise the training of a reference-based IQA metric.
The pointwise labels, which support universal comparison between arbitrary image pairs, enable an information-rich training objective.
Extensive experiments across diverse backbone architectures confirm the effectiveness of our generation pipeline, outperforming human-annotated datasets in multiple benchmarks. Codes are
publicly available at:
\url{https://github.com/JHLew/FoMo}
\end{abstract}

%% file: 01_intro.tex
\section{Introduction}
Reference-based image quality assessment (IQA) metrics~\cite{dists,dreamsim,pieapp,lpips,ssim} aim to quantify the perceptual difference between a reference image and its distorted counterpart. In particular, they play a central role in diverse image restoration tasks, such as super-resolution and denoising, where the goal is to compute the distance between a restored image and a reference target image.
A reliable metric must therefore align closely with human perceptual judgments, not only distinguishing which of two images is closer to the reference, but also inducing a globally consistent ordering across diverse distortion types and severity levels.

To train such metrics, IQA datasets ~\cite{lpips, pieapp, kadid10k}have employed different forms of human supervision to approximate perceptual rankings.
Mean opinion score (MOS)~\cite{tid2013, kadid10k} is the most direct and standard approach to produce such globally ordered labels. Multiple observers independently rate each distorted image on an absolute quality scale; the mean score is used as ground truth, and its global structure naturally supports rank-correlation training objectives~\cite{tid2013}. However, reliable MOS collection is both expensive and fragile. Due to annotator bias and inter-session inconsistency, a large number of responses per image is required to suppress noise. KADID-10k~\cite{kadid10k}, one of the largest MOS-annotated reference-based IQA datasets, required 30 crowdsourced ratings per image from over 2,200 subjects to produce 10,125 distorted images derived from only 81 reference images.
Furthermore, labels are known to be inconsistent across datasets:
the same distorted image can receive substantially different scores across datasets, because no common perceptual reference point exists across them.~\cite{pieapp}

These difficulties have driven the field toward pairwise preference labeling. The two-alternative forced choice (2AFC) protocol, asking annotators which of two distorted images is more similar to a reference, is considerably more reliable than absolute rating, as relative judgments are less susceptible to individual-scale biases~\cite{pieapp, lpips}. BAPPS ~\cite{lpips} and PieAPP~\cite{pieapp} established 2AFC as the foundation for learning modern perceptual metrics, and both demonstrate that pairwise labels exhibit higher inter-annotator agreement than MOS under equivalent collection conditions. 2AFC has since become the dominant annotation paradigm for learning-based IQA.
Yet, pairwise preference labels carry a structural limitation: a set of binary pairwise outcomes does not directly encode a global ordering, and global ranking of images are never taken into account in training.
This tension between the practical tractability of pairwise labeling and the global ranking objective of IQA is a recognized open challenge~\cite{ranksmoothed, gmciqa}.
Ideally, one would have access to dense labels that directly support optimization toward global rank-correlation.
In practice, however, this remains infeasible at scale under human annotation:
collecting clean and consistent pointwise quality signals across thousands of images, while controlling for the annotator noise endemic to MOS, is prohibitively expensive.

In this paper, we propose to circumvent this bottleneck through an automated dataset generation pipeline grounded in the generative dynamics of diffusion models.~\cite{ddpm,scoresde}
Our key insight is that the generative dynamics of diffusion models provide a natural proxy for perceptual distance. During generation, coarse image structure is established in the early timesteps, while fine-grained details are resolved only in later timesteps. 
In this work, we define a fork as a controlled branching of the denoising process: two samples follow an identical trajectory up to a selected timestep and are then generated independently thereafter.
Images that fork early in the generation process share only coarse structure and are perceptually far apart, whereas images that fork late differ primarily in fine detail.
We use this forking moment, FoMo, as a reference-grounded distance label for training a reference-based IQA metric. 

To validate this intuition, we conduct empirical analyses using a controlled set of image pairs generated via forking moments. We verify that their induced perceptual ordering aligns with human judgments, providing empirical grounding for using diffusion dynamics as a perceptual proxy. Building on this validation, we propose a fully automated and reference-grounded data generation pipeline that derives pointwise perceptual distance labels from diffusion forking moments. This enables global comparison across arbitrary image pairs without any human annotation, crowdsourcing infrastructure, or inter-annotator reconciliation.

Since FoMo is a pointwise score, it directly supports training consistent global ranking rather than binary pairwise preference.
We extensively validate the effectiveness of our method across multiple benchmarks and diverse architectural backbones, from CNN-based models~\cite{alexnet} to Transformer-based models.~\cite{vaswani2017attention}
In particular, our method outperforms human-annotated approaches, including KADID-10k's MOS labels, showing that automated diffusion-based labels can surpass large-scale human annotation as a strong and reliable training signal.
These results demonstrate that the FoMo can serve as a scalable, annotation-free alternative to both MOS and pairwise human labeling paradigms.
Our key contributions are summarized as below:

\begin{itemize}
    \item We propose a fully automated and annotation-free data generation pipeline for reference-based IQA that derives pointwise perceptual distance labels from the forking moments of diffusion trajectories, eliminating the need for human annotation.
    \item We demonstrate that diffusion generative dynamics encode perceptual distance in a manner well aligned with human visual judgment, providing a scalable alternative to MOS and pairwise preference annotations.
    \item We show that FoMo supervision enables globally consistent ranking across distorted images, and training with a RankNet-style~\cite{ranknet} global objective improves reference-based IQA performance across diverse benchmarks and model architectures.
\end{itemize}

%% file: 03_preliminary.tex
\section{Preliminary}
\paragraph{Diffusion Models, Flow Matching, and Rectified Flows.}
Denoising diffusion probabilistic models (DDPM)~\cite{ddpm} define a forward Markov process that gradually corrupts a clean image $x_0$ by adding Gaussian noise over $T$ discrete timesteps, yielding a sequence of increasingly noisy images $x_1, x_2, \ldots, x_T$ where $x_T \sim \mathcal{N}(0, I)$.
A neural network is trained to reverse this process, iteratively denoising $x_T$ back to a clean sample $x_0$. Score-based generative models~\cite{scoresde} generalize this to a continuous-time stochastic differential equation (SDE) framework, unifying many discrete diffusion variants under a single formalism.
Flow matching~\cite{flowmatching} and rectified flows~\cite{rectifiedflow} instead parameterize the generative process as a probability flow ODE along linear interpolations between data and noise: $x_t = (1-t)x_0 + t\epsilon$, where $\epsilon \sim \mathcal{N}(0, I)$ and $t \in [0, 1]$.
The resulting trajectories are straighter and more sample-efficient, motivating their adoption in state-of-the-art models such as FLUX~\cite{flux}.

Despite differences in trajectory geometry and training formulation, all three families share the same fundamental structure: a forward process that progressively destroys image information from fine detail toward coarse structure, and a learned reverse process that recovers the image from a stochastically sampled intermediate state. FoMo is grounded in this shared structure. Throughout this paper, we describe our method in the continuous-time SDE framework, as it provides the most general formulation. Most experiments in this paper are conducted with FLUX, which uses a rectified flow formulation where forward corruption follows $x_t = (1-t)x_0 + t\epsilon$.

\paragraph{Perceptual Structure Along the Generative Trajectory}
A key premise of FoMo is that the generative trajectory encodes perceptual information in a structured, timestep-dependent manner. Choi et al.~\cite{p2weighting} provide a direct characterization: at low noise levels (high signal-to-noise ratio), the reverse process recovers imperceptible fine-grained details; at intermediate noise levels, it reconstructs perceptually rich and discriminative content such as object structure and texture; at high noise levels, it recovers only coarse global attributes such as color distribution. This stratification is not incidental, it is a structural consequence of the forward process, which destroys information roughly monotonically from high-frequency to low-frequency.

These observations directly motivate FoMo. If two images share a reverse trajectory up to $t$ and then diverge via independent re-sampling, the perceptual content preserved up to $t$ is shared between them, while content destroyed before $t$ is independently regenerated. A late divergence (small $t$, little corruption) leaves most perceptual detail intact, yielding a perceptually close pair. An early divergence (large $t$, heavy corruption) destroys most discriminative content before re-sampling, producing a substantially different pair.

\paragraph{Trajectory Divergence as a Structural Prior}
The use of intermediate trajectory states to induce structured variation in generated outputs has appeared across several independent lines of work, lending support to the generality of the forking moment.~\cite{sdedit, ldm, decatur2025reusing} Most directly, Decatur et al.~\cite{decatur2025reusing} demonstrate that when generating a collection of semantically related images, early denoising steps capture shared structure across similar prompts and need only be computed once; trajectories then branch independently from a later timestep onward. This explicitly instantiates a tree-structured forking process, and their findings confirm that the branching timestep controls the degree of visual similarity among the resulting outputs. FoMo builds on this foundation by converting the forking structure into an explicit, scalable source of perceptual distance labels, replacing human annotation with the generative process itself.

\section{Empirical Grounding}
\label{sec:emp_grounding}
\input{figures/human_labeling}
Central to our approach is the hypothesis that the point of divergence in the diffusion sampling trajectory can serve as a meaningful perceptual distance label between image pairs. Before formalizing this as a metric, we first evaluate whether the divergence point serves as a reliable proxy for human perceptual similarity. That is, whether images forked earlier in the denoising process are consistently perceived as less similar to the reference than those forked later.

\subsection{Single-Reference Validation}
\label{sec:single_reference_study}
\paragraph{Study Design} To verify whether the divergence timestep provides perceptually meaningful guidance, we conducted a human study examining whether variants that diverge later in the denoising trajectory are consistently perceived as more similar to a reference image than those branching at earlier timesteps.
For each reference image, we synthesized five variants by injecting Gaussian noise at five distinct timesteps and denoising from each of them, so that later injection timesteps correspond to smaller perturbations and higher expected perceptual similarity to the reference. Participants were shown a reference image alongside its five variants and were asked to rank the variants from most to least similar compared to the reference. The workflow of this study is illustrated in Fig.~\ref{fig:single_ref_human_study}.

\paragraph{Results} According to our experimental analysis, human perceptual judgments demonstrate strong alignment with the divergence timestep ordering, yielding a Spearman rank correlation of 0.970 across 30 participants and 1,982 responses collected over 190 reference images sampled from the ImageNet~\cite{imagenet} validation set. Given an inter-rater correlation of 0.960, this level of agreement suggests that observer judgments are both consistent and well-structured. Collectively, these results indicate that the diffusion forking timestep constitutes a reliable proxy for perceptual similarity, providing empirical grounding for its adoption as a distance label in the subsequent metric formulation.

\subsection{Cross-Reference Validation}
\label{sec:cross_reference_study}
\paragraph{Study Design}
The study above establishes that the forking timestep orders variants of a single reference consistently with human perception. A perceptual distance, however, should also be globally consistent: if an image pair is labeled to be closer than another, they should look closer, regardless of which reference image anchors each pair.
We therefore ran a second study in the strict two-alternative forced-choice~(2AFC) format. Each item shows two reference: variant pairs built from two distinct references and asks which pair contains the images more similar to each other.
Since each forking timestep is chosen before its variant is generated, the two timesteps alone determine which pair our label calls closer, and no human answer enters the label. How hard an item is depends on the gap between its two forking timesteps: a small gap means both pairs were forked at nearly the same point, so they are almost equally similar, whereas a large gap sets a barely altered pair against a heavily altered one. We constructed 250 items, stratified into five bins of 50 by this gap, used each reference in at most one item, randomized left/right placement per participant, and collected 5,713 responses from 28 participants. Example questions from this study are in Fig.~\ref{fig:cross_ref_human_study}.

\paragraph{Results}
Human choices agree with the ordering induced by the forking timesteps in $90.2\%$ of individual responses, with a Fleiss' $\kappa$~\cite{fleiss1971} of $0.82$ indicating almost perfect inter-rater reliability~\cite{landis1977}.
Because every item was judged by many participants, we can also ask what they concluded collectively rather than one response at a time: taking the majority answer for each item, the label agrees with the human consensus on $92.8\%$ of items.
Agreement rises monotonically with the gap: $64.3\%$ for gaps of 1--10 timesteps, then $90.0\%$ for gaps of 11--20, $97.0\%$ for 21--30, and $99.5\%$ and $99.8\%$ for 31--40 and 41--50.
Counting consensus rather than individual votes, the share of items whose majority answer matches the label runs $69.4\%$, $94.0\%$, $100\%$, $100\%$ and $100\%$ across the same five bins: beyond a gap of 20 timesteps, every item is decided the way the label predicts.
Where the label agrees with people least, people also agree least with one another: in the narrowest bin two randomly chosen participants give the same answer on only $75.4\%$ of items and just $20\%$ of items are decided unanimously ($\kappa = 0.50$), against $99.6\%$, $96\%$ and $\kappa = 0.99$ in the widest.
The forking timestep therefore induces a similarity ordering that holds across different reference images, breaking down only where the two labels are too close to call.

\paragraph{Validation at Scale}
Since it is extremely difficult to conduct these comparisons at scale by human annotation, we repeat the same test with established perceptual metrics standing in for the human observer.
LPIPS-Alex~\cite{lpips}, LPIPS-VGG~\cite{lpips}, DISTS~\cite{dists} and DreamSim~\cite{dreamsim} are all fitted to human judgments and widely used as proxies for them, which makes them a reasonable substitute here.
We take 20,000 reference–variant pairs from our generated data and measure the distance each metric assigns to every pair. Pooling them into a single ranked list means that, as in the study above, almost every comparison is between pairs built on different references.
Against that pooled ranking the forking label reaches a Spearman Rank Order Correlation Coefficient (SROCC) of 0.932 with LPIPS-Alex, 0.915 with LPIPS-VGG, 0.904 with DISTS and 0.904 with DreamSim.
The ranking departs from the label only where the raters also hesitated, at near-ties. Once the two forking timesteps differ by more than 20 steps, the metrics agree with the label over 99.4\% of the time.
A pooled correlation of this magnitude is attainable only if the label is comparable across reference images rather than merely monotone within each one. The forking timestep behaves as a globally consistent distance label, not just a per-reference ranking.

%% file: figures/human_labeling.tex
\begin{figure}[htbp]
  \centering
  \begin{subfigure}[b]{0.59\textwidth}
      \centering
      \includegraphics[width=\textwidth]{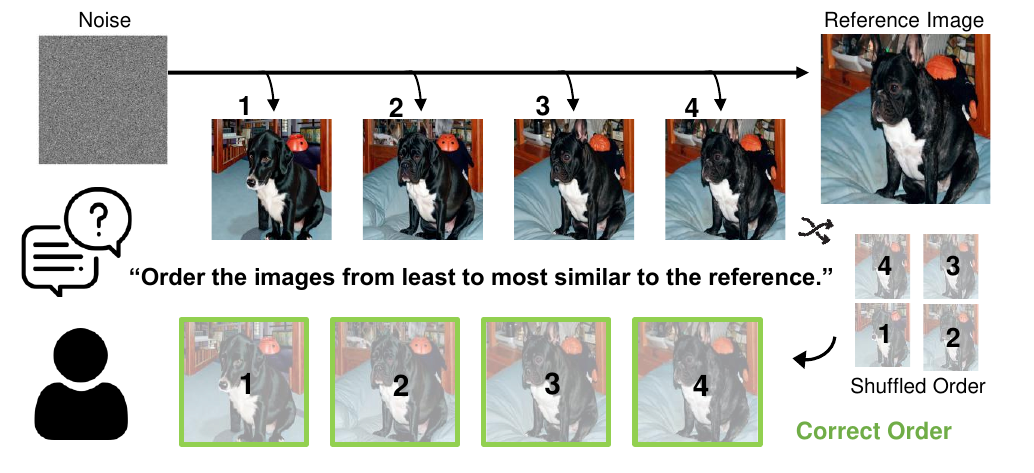}
      \caption{}
      \label{fig:single_ref_human_study}
  \end{subfigure}
  \hfill
  \hspace{-10pt}
  \begin{subfigure}[b]{0.42\textwidth}
      \centering
      \includegraphics[width=\textwidth]{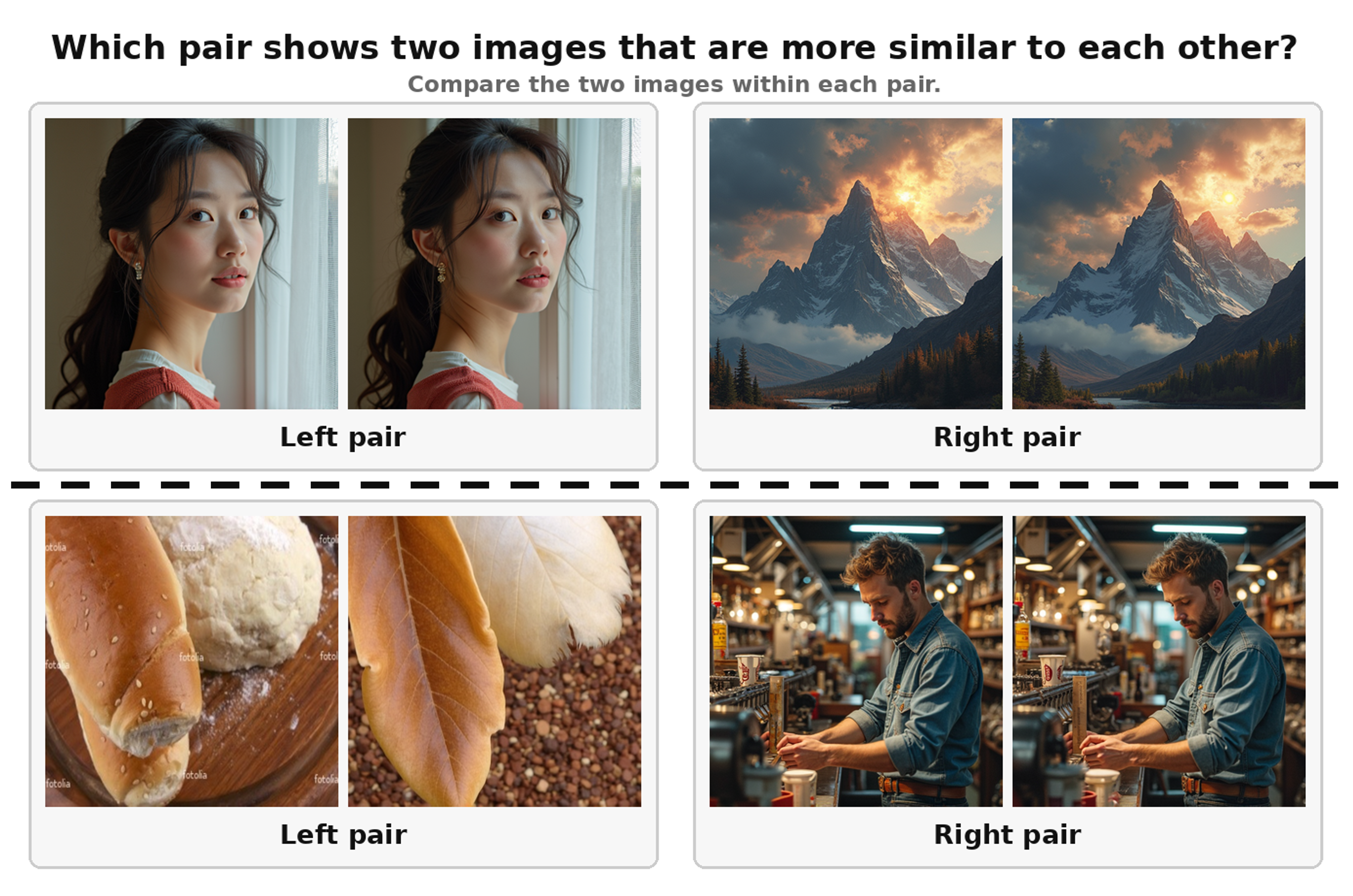}
      \caption{}
      \label{fig:cross_ref_human_study}
  \end{subfigure}
  \caption{
  (a) Single-reference study: participants rank synthesized variants to verify whether the divergence timestep in diffusion generative trajectory reflects the degree of perceptual similarity to the reference image. (b) Cross-reference study: participants are asked which of two pairs, built from different references, holds the two more similar images. Further examples in the Appendix.
  }
  \vspace{-5pt}
\end{figure}

%% file: 04_method.tex
\section{Method}
Based on the validation of Section~\ref{sec:emp_grounding}, we present our dataset generation pipeline with fully automated labeling, and detail the training procedure for learning a perceptual distance metric from it.
\input{figures/main_figure}

\subsection{Dataset Construction}
For data generation, we use FLUX.1-dev~\cite{flux} for its high-quality 
synthesis capability and broad coverage of visual content diversity.
Given a reference image $x_0$, we aim to generate a perturbed variant 
paired with a distance label that is precise and requires no human annotation.

Assume a denoising process with $S$ total steps. We uniformly sample a forking step $s \in [0, S-1]$, and obtain the corresponding interpolation factor from a predefined noise schedule $\mathcal{S}$, \textit{i.e.}, $t_s = \mathcal{S}(s)$. We then construct a noisy latent as
\begin{equation}
x_{t_s} = (1 - t_s)x_0 + t_s\epsilon, \quad \epsilon \sim \mathcal{N}(0, I).
\end{equation}

Starting from $x_{t_s}$, we perform the remaining $S-s$ denoising steps to obtain a perturbed image variant $x_0^{s}$. This yields a labeled pair $(x_0, x_0^{s}, t_s)$, where $t_s$ serves as the distance label. Intuitively, a larger $t_s$ corresponds to a higher noise level at the forking point and therefore to a greater perceptual deviation from the reference image.
Although stochasticity is inherent to the diffusion process, the automated nature of label generation enables large-scale sampling, reducing label variance and leading to stable convergence (See Sec.~\ref{app:variance} of Appendix.).
Data samples from our constructed dataset provided in Fig.~\ref{fig:samples}.

\subsection{Objective Function}
\label{sec:objective}
Conventional perceptual metrics such as LPIPS are trained on 
human-annotated 2AFC datasets, where each label encodes a relative 
preference between two distorted images given a shared reference. 
This relative structure constrains the loss to triplet-wise comparisons: given a triplet $(x_0, x_0', x_0'')$, binary cross-entropy is applied to the predicted probability that one variant is closer to the reference than the other.

Our labels, by contrast, are pointwise: each pair $(x_0, x_0^{s})$ 
carries an independent distance value $t_s$, without requiring a shared 
anchor for comparison. This permits a more expressive training objective. 
Specifically, for a batch of $B$ pairs with predicted distances 
$\{\hat{d}_i\}_{i=1}^B$ and labels $\{t_s^i\}_{i=1}^B$, we define a 
ground-truth comparison matrix $Y=[y_{ij}]$ as:
\begin{equation}
     y_{ij} = \begin{cases}
    \mathds{1}[t_s^i < t_s^j] & t_s^i \neq t_s^j, \quad i, j \in \{1, \ldots, B\} \\
    0.5 & \text{otherwise}
\end{cases}
\end{equation}

where $y_{ij} = 1$ indicates that pair $i$ has a smaller true distance 
than pair $j$.
We then apply binary cross-entropy loss over all $B \times B$ comparisons:
\begin{equation}
    \mathcal{L} = -\frac{1}{B^2} \sum_{i=1}^{B} \sum_{j=1}^{B}
    \Big[ y_{ij} \log \sigma(\hat{d}_j - \hat{d}_i)
    + \texttt{sg}((1 - y_{ij}) \log \sigma(\hat{d}_i - \hat{d}_j)) \Big],
\end{equation}
where $\sigma(\cdot)$ denotes the sigmoid function, and \texttt{sg} stands for stop-gradient operation.
This formulation, originally proposed for learning-to-rank in RankNet~\cite{ranknet}, is here adapted to perceptual distance learning, supervising the global ordering of distances across all $B \times B$ pairs rather than within isolated triplets. It makes full use of the pointwise label system from our data generation process.

%% file: figures/main_figure.tex
\begin{figure}
  \centering
  \includegraphics[width=\linewidth]{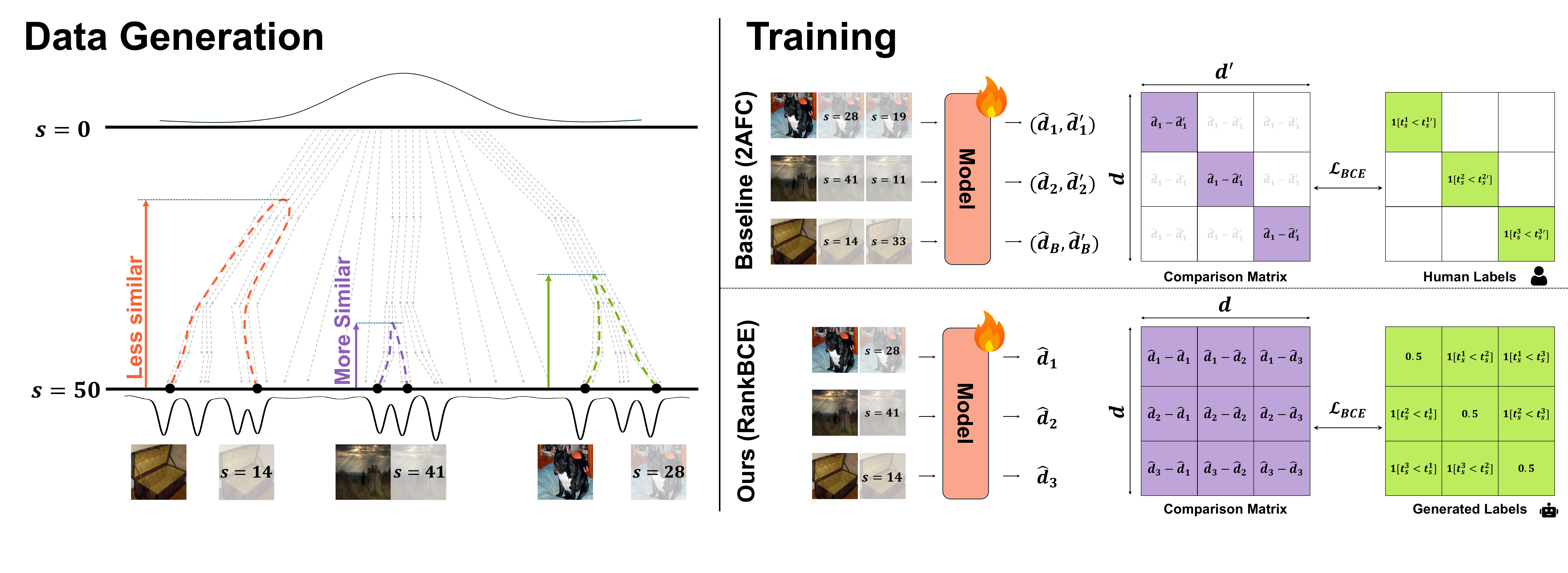}
  \vspace{-25pt}
  \caption{Visualization of our overall framework. (Left) Data generation trajectories, where the divergence point of the denoising trajectory (solid arrows) serves as a perceptual distance proxy. (Right) Unlike the traditional 2AFC framework, limited to within-anchor comparisons, our pointwise scoring system enables inter-anchor comparisons.}
  \vspace{-5pt}
  \label{fig:main_figure}
\end{figure}

%% file: 05_experiments.tex
\section{Experiments}
\input{figures/visualize}

\subsection{Experimental Settings}
We use a maximum of $S=50$ sampling steps with FLUX, and the forking moment is sampled from a uniform distribution: $s \sim U[0,49]$.
For training, we generate a total of 480k pairs and labels. Of these, 240k pairs use real images sampled from the ImageNet database~\cite{imagenet} as references, and the other 240k pairs use synthetic images generated by FLUX as references.
We mix these two domains of reference images in order to ensure a good coverage of real-world and synthetic domains.

We experiment both on models with CNN backbones and Transformer backbones.
For CNN backbones, we experiment with the well-established LPIPS~\cite{lpips} and DISTS~\cite{dists} backbones.
For Transformer backbones, we base upon the strong foundational models, DINOv3~\cite{dinov3}, CLIP~\cite{clip} and MAE~\cite{mae}, along with a recent DreamSim~\cite{dreamsim} backbone. 
Following the experimental setting from Zhang et al.~\cite{lpips}, we use the feature extractors fixed to the pre-trained state, and train the prediction head from scratch, except for DreamSim, which tunes the feature extractor with LoRA~\cite{lora} in its specific setting.
All evaluations are conducted under the native resolution of each image, except for DreamSim, which enforces 224 resolution images by resizing the input at all times.
All experiments reported in this paper were trained with models with approximately 240k iterations of updates, using a single NVIDIA RTX A40 GPU.
All quantitative results in the main manuscript is reported by Spearman Rank Order Correlation Coefficient (SROCC), unless otherwise mentioned.
Further details on experimental setting are described in the Appendix.

\subsection{Quantitative Evaluations}
\input{main_table}

The main quantitative result of our approach, in comparison to existing dataset and objectives, is presented in Table~\ref{tab:combined_results}.
Across four benchmarks and seven backbone architectures, FoMo achieves the best overall performance.
On PIPAL~\cite{pipal}, the least saturated and the most important benchmark, FoMo outperforms the existing datasets and their objectives, in all seven backbones by a large margin, especially in Transformer-based backbones.
On the other three benchmarks, TID2013~\cite{tid2013}, CSIQ~\cite{csiq} and LIVE~\cite{live}, FoMo performs comparable to baselines on CNN backbones, and outperforms most of them in Transformer backbones.
These results reflect the effectiveness of our data generation pipeline and objective function, despite being the only approach that does not require human annotation.
The entire numbers containing Kendall rank-order correlation coefficient (KROCC) and Pearson linear correlation coefficient (PLCC) on these benchmarks can be found in the Appendix.

\subsection{Ablation study}
In this section, we ablate and analyze the experimental choices in our pipeline.
First, we discuss on the experimental analysis on the label and objective function used in training.
Second, as our method is based on a RankNet-style binary cross-entropy loss which incorporates the entire ranking within a batch, computing a $B\times B$ comparison matrix, the batch size is expected to be an important factor in our experiments, and we study on its effects.
Third, we study to verify the generalizability of our approach, and check if the method could work with other diffusion models besides FLUX.
Fourth, we study which timesteps are important in training, by dropping certain timestep ranges in training. Unless otherwise mentioned, we experiment on LPIPS-Alex and DINOv3 backbones, as representatives of each backbone styles, CNNs and Transformers. All the experiments in this section is evaluated with PIPAL benchmark, with SROCC scores.

\paragraph{Objective and label type}
We study on the choices on the objective function and the label types. In this experiment, we also experiment on KADID-10k dataset, for comprehensive analysis.
First, our pipeline has two possible candidates of labels, using the interpolation factor $t$ or the forking timestep $s$ as the label, whereas KADID-10k labels are DMOS scores.
The main difference with using $t$ and step $s$ as the label is resolution-variance.
The main difference between using $t$ and $s$ as labels lies in their sensitivity to resolution changes, whether the label values are resolution-dependent.
Refer to Sec.~\ref{sec:random_resize_crop} of Appendix for further explanations.
For each point-wise labels, we can apply three forms of objective function in training, ranked binary cross-entropy (RankBCE) loss, 2AFC-style pairwise binary cross-entropy loss, and a simple L1 regression loss on the label value.
Our experimental results are presented in Table~\ref{tab:ablation_pred_type}.
According to our experiments, ranked binary cross-entropy loss as in RankNet~\cite{ranknet} has clearly proved to be consistently superior than other objectives,
supporting our claim on the importance of using rank-based objective.
While direct DMOS regression is the default training convention~\cite{kadid10k, dists} for KADID-10k, our results show that a rank-based BCE objective is the stronger choice, and under this objective, FoMo still remains the better training source.

\input{ablation_pred_type}

\paragraph{Analysis on Batch Size}
We study on the effects of batch size in our ranked binary cross-entropy loss. Pairwise ranking losses that operate over all within-batch pairs are known to benefit from large batch sizes, since a larger batch means more comparisons and thus richer supervision per training step, a property well-documented in contrastive learning~\cite{simclr, clip}.
As shown in Table~\ref{tab:ablation_batch_size}, we observe a similar trend. Performance improves with batch size increases, especially in Transformer-based DINOv3 backbone, whereas in CNN-based backbone LPIPS-Alex, the performance peaks at batch size of 64.
This reflects the potential that our approach could be scaled more with larger batch size and larger models.

\input{ablation_batch_size}

\paragraph{Cross Model Validation}
To study the generalizability of our method, we adopt diverse diffusion models into our pipeline in replacement to FLUX, which we used in our main experiments.
For this experiment, we use three pre-trained diffusion models publicly available, \textit{i.e.}, Stable Diffusion 1.5 (SD 1.5)~\cite{ldm}, SD-XL~\cite{sdxl}, and SD3~\cite{sd3}.
Using these pre-trained models, we generate image pair sets in the same manner.
For this experiment we build a pool of 50k pairs per generator and draw five
disjoint 10k subsets from it, training on each subset with the training seed
held fixed.
Every generator is thus matched at 10k training pairs, roughly $2\%$ of our main training set.
The experimental results can be found in Table~\ref{tab:ablation_sd}.
Across generators the ordering of the two backbones is preserved, and every generator yields a working metric. Even at that reduced budget, training on any of the four surpasses the best human-annotated dataset for the CNN backbone ($0.622$, Table~\ref{tab:combined_results}), and with the Transformer backbone every generator except SD-XL at least matches its best human-annotated baseline ($0.508$). The approach therefore does not depend on FLUX specifically, but the choice of generator is not immaterial either. FLUX.1 is the strongest option for both backbones, by margins well outside the data-sampling error bars, and the Transformer-based models benefit the most from the strong generator.
\label{sec:ablation}
\input{ablation_sd}

\paragraph{Timestep Range Selection}
We study on how the performance of models change depending on the sampling range of forking timesteps. In this experiment, we use the three CNN backbones, LPIPS-Alex, VGG and DISTS models, along with a Transformer backbone, DINOv3. The results are displayed in Table~\ref{tab:ablation_timestep}. Within a 50-step process, the initial 35 steps, which are closer to the noise state than the clean image state, has shown to be the most important. The final 15 steps of the generation process are dedicated to refining fine-grained details that are barely perceptible to the human eye. Since such subtle refinements carry little perceptual significance, incorporating these steps into training may introduce noise into the learning signal, and the two LPIPS backbones indeed peak without them. DISTS and DINOv3, however, perform best over the full range. Considering the overall robustness, in our main experiments, we used the full 50 steps in training.

\input{ablation_timestep}

%% file: figures/visualize.tex
\begin{figure}
  \centering
  \includegraphics[width=\linewidth]{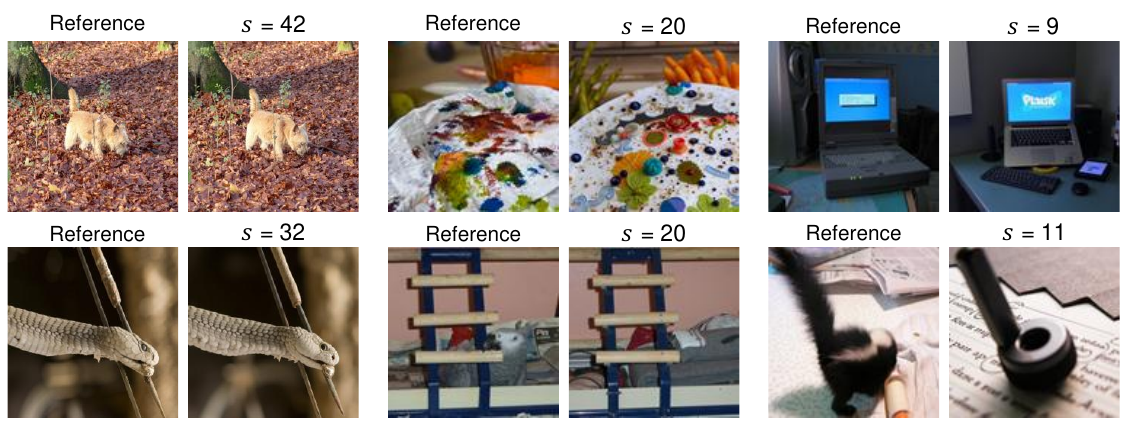}
  \caption{Data sample image pairs and their labels from our constructed dataset. $s$ denote the forking step (smaller-the-further), or the distance label of the image with respect to the reference image.}
  \label{fig:samples}
  \vspace{-5pt}
\end{figure}

%% file: main_table.tex
\begin{table*}[htbp]
    \centering
    \caption{Comparison of training datasets and their objectives on four reference-based IQA benchmarks, across CNN- and Transformer-based backbones. Mean SROCC over five random seeds; standard deviations are given in Table~\ref{tab:combined_results_srocc_full}. $^\dagger$ Evaluated under 224 resolution.}
    \resizebox{\textwidth}{!}{
    \begin{tabular}{l | c | c |ccc | cccc | c}
    \toprule
        \multirow{2.5}{*}{\textbf{Train Data}} & \multirow{2.5}{*}{\makecell{\textbf{Human} \\ \textbf{Annot.}}} & \multirow{2.5}{*}{\textbf{Objective}}
        &\multicolumn{3}{c|}{\textbf{CNN-based}} & \multicolumn{4}{c|}{\textbf{Transformer-based}} & \multirow{2.5}{*}{\textbf{Avg.}}\\
        \cmidrule(lr){4-6} \cmidrule(lr){7-10}
        & & & LPIPS-Alex & LPIPS-VGG & DISTS & DINOv3 & CLIP & MAE & DreamSim$^\dagger$ \\
    \midrule
    \multicolumn{11}{c}{\textit{Results on PIPAL~\cite{pipal}}} \\
    \midrule
        BAPPS~\cite{lpips} & \cmark & 2AFC & \underline{0.622} & 0.634 & 0.489 & 0.287 & 0.151 & \underline{0.338} & \underline{0.760} & 0.469 \\
        PieAPP~\cite{pieapp} & \cmark & 2AFC & 0.602 & \underline{0.635} & \underline{0.564} & \underline{0.325} & 0.258 & 0.294 & 0.711 & \underline{0.484} \\
        NIGHTS~\cite{dreamsim} & \cmark & 2AFC & 0.577 & 0.545 & -0.294 & 0.234 & 0.277 & 0.234 & 0.662 & 0.319 \\
        KADID-10K~\cite{kadid10k} & \cmark & L1 & 0.577 & 0.590 & 0.518 & 0.298 & \underline{0.312} & 0.253 & 0.617 & 0.452 \\
        FoMo (\textbf{Ours}) & \xmark & RankBCE & \textbf{0.733} & \textbf{0.683} & \textbf{0.615} & \textbf{0.699} & \textbf{0.644} & \textbf{0.632} & \textbf{0.776} & \textbf{0.683} \\
    \midrule
    \multicolumn{11}{c}{\textit{Results on TID2013~\cite{tid2013}}} \\
    \midrule
        BAPPS~\cite{lpips} & \cmark & 2AFC & 0.779 & 0.671 & 0.632 & 0.341 & 0.285 & 0.287 & \textbf{0.813} & 0.544 \\
        PieAPP~\cite{pieapp} & \cmark & 2AFC & 0.761 & \underline{0.722} & 0.680 & 0.315 & 0.302 & 0.467 & 0.767 & 0.573 \\
        NIGHTS~\cite{dreamsim} & \cmark & 2AFC & 0.779 & 0.653 & -0.497 & 0.230 & 0.294 & 0.358 & 0.762 & 0.368 \\
        KADID-10K~\cite{kadid10k} & \cmark & L1 & \textbf{0.793} & \textbf{0.757} & \textbf{0.834} & \underline{0.613} & \underline{0.519} & \underline{0.595} & 0.788 & \underline{0.700} \\
        FoMo (\textbf{Ours}) & \xmark & RankBCE & \underline{0.785} & 0.663 & \underline{0.691} & \textbf{0.713} & \textbf{0.737} & \textbf{0.644} & \underline{0.801} & \textbf{0.719} \\
    \midrule
    \multicolumn{11}{c}{\textit{Results on CSIQ~\cite{csiq}}} \\
    \midrule
        BAPPS~\cite{lpips} & \cmark & 2AFC & \underline{0.943} & 0.887 & 0.855 & 0.411 & 0.427 & 0.455 & \textbf{0.911} & 0.699 \\
        PieAPP~\cite{pieapp} & \cmark & 2AFC & 0.936 & \textbf{0.896} & 0.821 & 0.331 & 0.514 & 0.587 & \underline{0.903} & 0.713 \\
        NIGHTS~\cite{dreamsim} & \cmark & 2AFC & \textbf{0.945} & 0.862 & -0.560 & 0.335 & 0.364 & 0.427 & 0.867 & 0.463 \\
        KADID-10K~\cite{kadid10k} & \cmark & L1 & 0.935 & \underline{0.895} & \textbf{0.938} & \underline{0.616} & \underline{0.604} & \underline{0.698} & 0.874 & \underline{0.794} \\
        FoMo (\textbf{Ours}) & \xmark & RankBCE & 0.938 & 0.859 & \underline{0.918} & \textbf{0.811} & \textbf{0.901} & \textbf{0.795} & 0.894 & \textbf{0.874} \\
    \midrule
    \multicolumn{11}{c}{\textit{Results on LIVE~\cite{live}}} \\
    \midrule
        BAPPS~\cite{lpips} & \cmark & 2AFC & \textbf{0.952} & \textbf{0.929} & 0.843 & 0.688 & 0.425 & 0.298 & \underline{0.927} & 0.723 \\
        PieAPP~\cite{pieapp} & \cmark & 2AFC & 0.934 & 0.921 & 0.860 & 0.458 & 0.415 & 0.537 & 0.906 & 0.719 \\
        NIGHTS~\cite{dreamsim} & \cmark & 2AFC & \underline{0.949} & 0.919 & -0.774 & 0.439 & 0.560 & 0.376 & 0.867 & 0.477 \\
        KADID-10K~\cite{kadid10k} & \cmark & L1 & 0.942 & 0.912 & \underline{0.950} & \underline{0.765} & \underline{0.827} & \underline{0.797} & 0.896 & \underline{0.870} \\
        FoMo (\textbf{Ours}) & \xmark & RankBCE & 0.948 & \underline{0.923} & \textbf{0.954} & \textbf{0.896} & \textbf{0.925} & \textbf{0.907} & \textbf{0.931} & \textbf{0.926} \\
    \bottomrule
    \end{tabular}
    }
    \vspace{-5pt}
    \label{tab:combined_results}
\end{table*}

%% file: ablation_pred_type.tex
\begin{table*}[htbp]
    \centering
    \caption{Comparative experiment on the objective function and label type. Ranked binary cross-entropy loss (Rank) consistently shows stronger results, compared to triplet-based paired cross-entropy loss (2AFC) and L1 regression loss on the label (L1). Our label has shown to be more helpful in training, compared to the large-scale human annotated KADID-10k. Mean over five random seeds; full numbers with standard deviations can be found in Table~\ref{tab:ablation_pred_type_full}.}
    \resizebox{\textwidth}{!}{
    \begin{tabular}{l | c c c | c c c | c c c}
    \toprule
        & \multicolumn{6}{c|}{\textbf{FoMo (Ours)}} & \multicolumn{3}{c}{\textbf{KADID-10K~\cite{kadid10k}}} \\
        \cmidrule(lr){2-7} \cmidrule(lr){8-10}
        \textbf{Label} & \multicolumn{3}{c|}{$t$} & \multicolumn{3}{c|}{$s$} & \multicolumn{3}{c}{DMOS} \\
        \cmidrule(lr){2-4} \cmidrule(lr){5-7} \cmidrule(lr){8-10}
        \textbf{Objective} & Rank & 2AFC & L1 & Rank & 2AFC & L1 & Rank & 2AFC & L1 \\
    \midrule
        LPIPS-Alex~\cite{lpips} & \textbf{0.733} & \underline{0.592} & 0.440 & \textbf{0.733} & \underline{0.592} & 0.390 & \textbf{0.625} & \underline{0.611} & 0.577 \\
        DINOv3~\cite{dinov3} & \textbf{0.699} & 0.622 & \underline{0.694} & \textbf{0.697} & 0.614 & \underline{0.695} & \textbf{0.309} & 0.145 & \underline{0.298} \\
    \bottomrule
    \end{tabular}
    }
    \label{tab:ablation_pred_type}
\end{table*}

%% file: ablation_batch_size.tex
\begin{table}[htbp]
    \centering
    \caption{Effect of batch size on training performance. The Transformer-based backbone improves monotonically with batch size, while the CNN-based backbone saturates at 64 and degrades slightly beyond it. Mean $\pm$ standard deviation over five random seeds.}
    \resizebox{\columnwidth}{!}{
    \begin{tabular}{l | c c c c c}
    \toprule
        \multirow{2.5}{*}{\textbf{Backbone}} & \multicolumn{5}{c}{\textbf{Batch size}} \\
        & 16 & 32 & 64 & 128 & 256 \\
    \midrule
        LPIPS-Alex~\cite{lpips} & 0.719 $\pm$ 0.006 & 0.729 $\pm$ 0.006 & \textbf{0.733 $\pm$ 0.006} & 0.728 $\pm$ 0.006 & 0.712 $\pm$ 0.006 \\
        DINOv3~\cite{dinov3} & 0.479 $\pm$ 0.080 & 0.650 $\pm$ 0.018 & 0.696 $\pm$ 0.007 & 0.705 $\pm$ 0.003 & \textbf{0.707 $\pm$ 0.005} \\
    \bottomrule
    \end{tabular}
    }
    \vspace{-5pt}
    \label{tab:ablation_batch_size}
\end{table}

%% file: ablation_sd.tex
\begin{table}[htbp]
    \centering
    \caption{Experimental results using diverse diffusion models for data generation. Our pipeline is not specific to FLUX, and every generator yields a working metric, although FLUX.1 is the strongest choice. Cells report mean $\pm$ standard deviation over five train runs, each using disjoint 10k subsets from the generator.}
    \begin{tabular}{l | c c c c}
    \toprule
        \multirow{2.5}{*}{\textbf{Backbone}} & \multicolumn{4}{c}{\textbf{Data Generator} (10k pairs)} \\
        & SD-1.5~\cite{ldm} & SD-XL~\cite{sdxl} & SD-3~\cite{sd3} & FLUX.1~\cite{flux} \\
    \midrule
        LPIPS-Alex~\cite{lpips} & 0.687 $\pm$ 0.008 & 0.662 $\pm$ 0.008 & 0.695 $\pm$ 0.007 & 0.741 $\pm$ 0.002 \\
        DINOv3~\cite{dinov3} & 0.574 $\pm$ 0.078 & 0.418 $\pm$ 0.044 & 0.517 $\pm$ 0.017 & 0.672 $\pm$ 0.016 \\
    \bottomrule
    \end{tabular}
    \label{tab:ablation_sd}
\end{table}

%% file: ablation_timestep.tex
\begin{table}[htbp]
    \centering
    \caption{Experiment on different sampling ranges of forking timesteps. At a matched window width the earlier window is the most useful one, and the full schedule overall provides the most stable result. Mean $\pm$ standard deviation over five random seeds.}
    \resizebox{\columnwidth}{!}{
    \begin{tabular}{l | c | c c c c c}
    \toprule
        \multirow{2.5}{*}{\textbf{Backbone}} & \textbf{All} & \multicolumn{5}{c}{\textbf{Sampling Range of Forking Timesteps}} \\
        & $[0, 50]$ & $[0, 25]$ & $[0, 35]$ & $[13, 37]$ & $[15, 50]$ & $[25, 50]$ \\
    \midrule
        LPIPS-Alex~\cite{lpips} & 0.733 $\pm$ 0.006 & \underline{0.748 $\pm$ 0.008} & \textbf{0.757 $\pm$ 0.007} & 0.729 $\pm$ 0.008 & 0.731 $\pm$ 0.007 & 0.720 $\pm$ 0.006 \\
        LPIPS-VGG~\cite{lpips} & 0.683 $\pm$ 0.008 & \underline{0.688 $\pm$ 0.005} & \textbf{0.697 $\pm$ 0.006} & 0.685 $\pm$ 0.004 & 0.666 $\pm$ 0.006 & 0.626 $\pm$ 0.005 \\
        DISTS~\cite{dists} & \textbf{0.615 $\pm$ 0.001} & 0.464 $\pm$ 0.002 & \underline{0.583 $\pm$ 0.001} & 0.569 $\pm$ 0.002 & 0.556 $\pm$ 0.001 & 0.485 $\pm$ 0.003 \\
        DINOv3~\cite{dinov3} & \textbf{0.703 $\pm$ 0.006} & 0.620 $\pm$ 0.015 & 0.696 $\pm$ 0.015 & 0.700 $\pm$ 0.009 & \underline{0.702 $\pm$ 0.015} & 0.680 $\pm$ 0.014 \\
    \bottomrule
    \end{tabular}
    }
    \label{tab:ablation_timestep}
\end{table}

%% file: 06_conclusion.tex
\section{Conclusion}
In this paper, we introduced a data generation pipeline that reframes perceptual distance as a forking moment in a diffusion denoising trajectory.
By forward-diffusing a reference image to a sampled timestep and denoising it back, we automatically synthesize training pairs with calibrated perceptual distances, no human annotation required.
Our human study verified that using forking moments in diffusion trajectory aligns well with human perception, supporting our argument.
We further show that RankNet-style supervision over our generated data substantially outperforms 2AFC-style binary classification, yielding richer gradient signal and implicit transitivity enforcement. Extensive experiments on diverse backbones and evaluation benchmarks demonstrate consistent improvements over strong baselines, validating both the dataset generation pipeline and the ranking-based objective.

%% file: appendix.tex
\appendix
\section{Experimental Details}
\paragraph{Model Training}
The prediction heads of LPIPS-Alex/VGG are linear layers from each level of the feature pyramid, and DISTS learn the $\alpha$ and $\beta$ in integration of the extracted intermediate features.
For three Transformer backbones, DINOv3, CLIP and MAE, the feature extractor is also fixed, and we append three ViT layers~\cite{vit} for distance logit prediction. The prediction head layers take the feature tokens extracted from the backbone as the input, along with a \texttt{[CLS]} token which make the prediction of the distance logit.
For all experiments besides DreamSim, we use a batch size of 64, and a fixed learning rate of 2e-4 and 1e-4 respectively for CNN backbones and the three Transformer backbones.
For DreamSim, we follow its original configuration without any change, from LoRA~\cite{lora} configuration ($r{=}16$, $\alpha{=}1$, dropout $0.3$ on the qkv projections), input protocol and optimizer, and only the training set is varied; the sample budget matches every other column.

\paragraph{Ensuring symmetry in Transformer backbone models}
We write $\hat{d}(u, v)$ for the distance the model assigns to an image pair $(u, v)$; this is the quantity written $\hat{d}_i$ in Sec.~\ref{sec:objective}. For the Transformer backbones, DINOv3, CLIP and MAE, it is produced by the prediction head, whose output needs an adjustment for symmetry. The head reads the two images as a single concatenated sequence and is thus asymmetric, $\hat{d}(u, v)$ and $\hat{d}(v, u)$ returning non-identical values. Therefore we report the results acquired from $\tfrac{1}{2}\big[\hat{d}(u, v) + \hat{d}(v, u)\big]$, which removes the asymmetry. During training, each training pair is presented in a random order to make the model work in both orders, and be naturally symmetric. This way, we observe the distance logits to be similar in both input orders, although inherently they cannot be perfectly identical. Neither adjustment applies to LPIPS, DISTS or DreamSim, which compute a symmetric distance directly and are reported as they are.

\section{Full Results}


\begin{table*}[htbp]
    \centering
    \caption{Comparison of training datasets and their objectives on four reference-based IQA benchmarks, across CNN- and Transformer-based backbones. Mean $\pm$ standard deviation of SROCC over five random seeds; full-statistics version of Table~\ref{tab:combined_results}. $^\dagger$Evaluated under 224 resolution.}
    \vspace{5pt}
    \resizebox{\textwidth}{!}{
    \begin{tabular}{l |ccc | cccc | c}
    \toprule
        \multirow{2.5}{*}{\textbf{Train Data}}
        &\multicolumn{3}{c|}{\textbf{CNN-based}} & \multicolumn{4}{c|}{\textbf{Transformer-based}} & \multirow{2.5}{*}{\textbf{Avg.}}\\
        \cmidrule(lr){2-4} \cmidrule(lr){5-8}
        & LPIPS-Alex & LPIPS-VGG & DISTS & DINOv3 & CLIP & MAE & DreamSim$^\dagger$ \\
    \midrule
    \multicolumn{9}{c}{\textit{Results on PIPAL~\cite{pipal}}} \\
    \midrule
        BAPPS~\cite{lpips} & \underline{0.622 $\pm$ 0.005} & 0.634 $\pm$ 0.018 & 0.489 $\pm$ 0.003 & 0.287 $\pm$ 0.020 & 0.151 $\pm$ 0.040 & \underline{0.338 $\pm$ 0.013} & \underline{0.760 $\pm$ 0.007} & 0.469 $\pm$ 0.008 \\
        PieAPP~\cite{pieapp} & 0.602 $\pm$ 0.008 & \underline{0.635 $\pm$ 0.016} & \underline{0.564 $\pm$ 0.003} & \underline{0.325 $\pm$ 0.017} & 0.258 $\pm$ 0.037 & 0.294 $\pm$ 0.023 & 0.711 $\pm$ 0.007 & \underline{0.484 $\pm$ 0.007} \\
        NIGHTS~\cite{dreamsim} & 0.577 $\pm$ 0.006 & 0.545 $\pm$ 0.011 & -0.294 $\pm$ 0.003 & 0.234 $\pm$ 0.024 & 0.277 $\pm$ 0.025 & 0.234 $\pm$ 0.029 & 0.662 $\pm$ 0.019 & 0.319 $\pm$ 0.005 \\
        KADID-10K~\cite{kadid10k} & 0.577 $\pm$ 0.025 & 0.590 $\pm$ 0.034 & 0.518 $\pm$ 0.002 & 0.298 $\pm$ 0.010 & \underline{0.312 $\pm$ 0.018} & 0.253 $\pm$ 0.041 & 0.617 $\pm$ 0.003 & 0.452 $\pm$ 0.006 \\
        FoMo (\textbf{Ours}) & \textbf{0.733 $\pm$ 0.006} & \textbf{0.683 $\pm$ 0.008} & \textbf{0.615 $\pm$ 0.001} & \textbf{0.699 $\pm$ 0.006} & \textbf{0.644 $\pm$ 0.024} & \textbf{0.632 $\pm$ 0.020} & \textbf{0.776 $\pm$ 0.010} & \textbf{0.683 $\pm$ 0.004} \\
    \midrule
    \multicolumn{9}{c}{\textit{Results on TID2013~\cite{tid2013}}} \\
    \midrule
        BAPPS~\cite{lpips} & 0.779 $\pm$ 0.003 & 0.671 $\pm$ 0.002 & 0.632 $\pm$ 0.001 & 0.341 $\pm$ 0.022 & 0.285 $\pm$ 0.030 & 0.287 $\pm$ 0.022 & \textbf{0.813 $\pm$ 0.004} & 0.544 $\pm$ 0.006 \\
        PieAPP~\cite{pieapp} & 0.761 $\pm$ 0.006 & \underline{0.722 $\pm$ 0.005} & 0.680 $\pm$ 0.004 & 0.315 $\pm$ 0.042 & 0.302 $\pm$ 0.025 & 0.467 $\pm$ 0.027 & 0.767 $\pm$ 0.004 & 0.573 $\pm$ 0.008 \\
        NIGHTS~\cite{dreamsim} & 0.779 $\pm$ 0.003 & 0.653 $\pm$ 0.001 & -0.497 $\pm$ 0.005 & 0.230 $\pm$ 0.022 & 0.294 $\pm$ 0.035 & 0.358 $\pm$ 0.033 & 0.762 $\pm$ 0.007 & 0.368 $\pm$ 0.007 \\
        KADID-10K~\cite{kadid10k} & \textbf{0.793 $\pm$ 0.004} & \textbf{0.757 $\pm$ 0.012} & \textbf{0.834 $\pm$ 0.001} & \underline{0.613 $\pm$ 0.024} & \underline{0.519 $\pm$ 0.016} & \underline{0.595 $\pm$ 0.013} & 0.788 $\pm$ 0.004 & \underline{0.700 $\pm$ 0.006} \\
        FoMo (\textbf{Ours}) & \underline{0.785 $\pm$ 0.003} & 0.663 $\pm$ 0.003 & \underline{0.691 $\pm$ 0.001} & \textbf{0.713 $\pm$ 0.003} & \textbf{0.737 $\pm$ 0.016} & \textbf{0.644 $\pm$ 0.052} & \underline{0.801 $\pm$ 0.003} & \textbf{0.719 $\pm$ 0.007} \\
    \midrule
    \multicolumn{9}{c}{\textit{Results on CSIQ~\cite{csiq}}} \\
    \midrule
        BAPPS~\cite{lpips} & \underline{0.943 $\pm$ 0.001} & 0.887 $\pm$ 0.001 & 0.855 $\pm$ 0.001 & 0.411 $\pm$ 0.027 & 0.427 $\pm$ 0.015 & 0.455 $\pm$ 0.051 & \textbf{0.911 $\pm$ 0.000} & 0.699 $\pm$ 0.007 \\
        PieAPP~\cite{pieapp} & 0.936 $\pm$ 0.002 & \textbf{0.896 $\pm$ 0.007} & 0.821 $\pm$ 0.003 & 0.331 $\pm$ 0.042 & 0.514 $\pm$ 0.016 & 0.587 $\pm$ 0.012 & \underline{0.903 $\pm$ 0.003} & 0.713 $\pm$ 0.005 \\
        NIGHTS~\cite{dreamsim} & \textbf{0.945 $\pm$ 0.001} & 0.862 $\pm$ 0.002 & -0.560 $\pm$ 0.006 & 0.335 $\pm$ 0.031 & 0.364 $\pm$ 0.022 & 0.427 $\pm$ 0.033 & 0.867 $\pm$ 0.007 & 0.463 $\pm$ 0.006 \\
        KADID-10K~\cite{kadid10k} & 0.935 $\pm$ 0.002 & \underline{0.895 $\pm$ 0.016} & \textbf{0.938 $\pm$ 0.000} & \underline{0.616 $\pm$ 0.009} & \underline{0.604 $\pm$ 0.026} & \underline{0.698 $\pm$ 0.035} & 0.874 $\pm$ 0.009 & \underline{0.794 $\pm$ 0.010} \\
        FoMo (\textbf{Ours}) & 0.938 $\pm$ 0.001 & 0.859 $\pm$ 0.006 & \underline{0.918 $\pm$ 0.001} & \textbf{0.811 $\pm$ 0.002} & \textbf{0.901 $\pm$ 0.014} & \textbf{0.795 $\pm$ 0.038} & 0.894 $\pm$ 0.002 & \textbf{0.874 $\pm$ 0.005} \\
    \midrule
    \multicolumn{9}{c}{\textit{Results on LIVE~\cite{live}}} \\
    \midrule
        BAPPS~\cite{lpips} & \textbf{0.952 $\pm$ 0.001} & \textbf{0.929 $\pm$ 0.002} & 0.843 $\pm$ 0.001 & 0.688 $\pm$ 0.032 & 0.425 $\pm$ 0.059 & 0.298 $\pm$ 0.062 & \underline{0.927 $\pm$ 0.001} & 0.723 $\pm$ 0.013 \\
        PieAPP~\cite{pieapp} & 0.934 $\pm$ 0.013 & 0.921 $\pm$ 0.013 & 0.860 $\pm$ 0.002 & 0.458 $\pm$ 0.022 & 0.415 $\pm$ 0.030 & 0.537 $\pm$ 0.016 & 0.906 $\pm$ 0.004 & 0.719 $\pm$ 0.006 \\
        NIGHTS~\cite{dreamsim} & \underline{0.949 $\pm$ 0.002} & 0.919 $\pm$ 0.002 & -0.774 $\pm$ 0.004 & 0.439 $\pm$ 0.021 & 0.560 $\pm$ 0.057 & 0.376 $\pm$ 0.058 & 0.867 $\pm$ 0.007 & 0.477 $\pm$ 0.015 \\
        KADID-10K~\cite{kadid10k} & 0.942 $\pm$ 0.013 & 0.912 $\pm$ 0.032 & \underline{0.950 $\pm$ 0.000} & \underline{0.765 $\pm$ 0.023} & \underline{0.827 $\pm$ 0.009} & \underline{0.797 $\pm$ 0.015} & 0.896 $\pm$ 0.012 & \underline{0.870 $\pm$ 0.012} \\
        FoMo (\textbf{Ours}) & 0.948 $\pm$ 0.003 & \underline{0.923 $\pm$ 0.004} & \textbf{0.954 $\pm$ 0.000} & \textbf{0.896 $\pm$ 0.002} & \textbf{0.925 $\pm$ 0.011} & \textbf{0.907 $\pm$ 0.026} & \textbf{0.931 $\pm$ 0.002} & \textbf{0.926 $\pm$ 0.003} \\
    \bottomrule
    \end{tabular}
    }
    \label{tab:combined_results_srocc_full}
\end{table*}


\begin{table*}[htbp]
    \centering
    \caption{Comparison of training datasets and their objectives on four reference-based IQA benchmarks, across CNN- and Transformer-based backbones. Mean $\pm$ standard deviation of KROCC over five random seeds. $^\dagger$Evaluated under 224 resolution.}
    \vspace{5pt}
    \resizebox{\textwidth}{!}{
    \begin{tabular}{l |ccc | cccc | c}
    \toprule
        \multirow{2.5}{*}{\textbf{Train Data}}
        &\multicolumn{3}{c|}{\textbf{CNN-based}} & \multicolumn{4}{c|}{\textbf{Transformer-based}} & \multirow{2.5}{*}{\textbf{Avg.}}\\
        \cmidrule(lr){2-4} \cmidrule(lr){5-8}
        & LPIPS-Alex & LPIPS-VGG & DISTS & DINOv3 & CLIP & MAE & DreamSim$^\dagger$ \\
    \midrule
    \multicolumn{9}{c}{\textit{Results on PIPAL~\cite{pipal}}} \\
    \midrule
        BAPPS~\cite{lpips} & \underline{0.440 $\pm$ 0.004} & \underline{0.455 $\pm$ 0.014} & 0.337 $\pm$ 0.002 & 0.194 $\pm$ 0.014 & 0.101 $\pm$ 0.027 & \underline{0.229 $\pm$ 0.009} & \underline{0.564 $\pm$ 0.007} & 0.332 $\pm$ 0.006 \\
        PieAPP~\cite{pieapp} & 0.420 $\pm$ 0.007 & 0.450 $\pm$ 0.013 & \underline{0.389 $\pm$ 0.002} & \underline{0.222 $\pm$ 0.013} & 0.175 $\pm$ 0.026 & 0.197 $\pm$ 0.016 & 0.511 $\pm$ 0.006 & \underline{0.338 $\pm$ 0.005} \\
        NIGHTS~\cite{dreamsim} & 0.405 $\pm$ 0.005 & 0.385 $\pm$ 0.008 & -0.204 $\pm$ 0.002 & 0.159 $\pm$ 0.017 & 0.186 $\pm$ 0.017 & 0.157 $\pm$ 0.020 & 0.471 $\pm$ 0.018 & 0.223 $\pm$ 0.004 \\
        KADID-10K~\cite{kadid10k} & 0.403 $\pm$ 0.020 & 0.419 $\pm$ 0.027 & 0.360 $\pm$ 0.001 & 0.204 $\pm$ 0.007 & \underline{0.212 $\pm$ 0.012} & 0.170 $\pm$ 0.028 & 0.440 $\pm$ 0.003 & 0.315 $\pm$ 0.005 \\
        FoMo (\textbf{Ours}) & \textbf{0.536 $\pm$ 0.006} & \textbf{0.497 $\pm$ 0.007} & \textbf{0.435 $\pm$ 0.001} & \textbf{0.500 $\pm$ 0.005} & \textbf{0.459 $\pm$ 0.019} & \textbf{0.444 $\pm$ 0.015} & \textbf{0.571 $\pm$ 0.010} & \textbf{0.492 $\pm$ 0.003} \\
    \midrule
    \multicolumn{9}{c}{\textit{Results on TID2013~\cite{tid2013}}} \\
    \midrule
        BAPPS~\cite{lpips} & 0.584 $\pm$ 0.003 & 0.497 $\pm$ 0.002 & 0.450 $\pm$ 0.001 & 0.235 $\pm$ 0.017 & 0.194 $\pm$ 0.021 & 0.192 $\pm$ 0.015 & \textbf{0.619 $\pm$ 0.003} & 0.396 $\pm$ 0.004 \\
        PieAPP~\cite{pieapp} & 0.567 $\pm$ 0.006 & \underline{0.540 $\pm$ 0.004} & 0.486 $\pm$ 0.004 & 0.217 $\pm$ 0.030 & 0.207 $\pm$ 0.017 & 0.321 $\pm$ 0.019 & 0.574 $\pm$ 0.003 & 0.416 $\pm$ 0.006 \\
        NIGHTS~\cite{dreamsim} & 0.582 $\pm$ 0.003 & 0.480 $\pm$ 0.001 & -0.345 $\pm$ 0.004 & 0.157 $\pm$ 0.016 & 0.199 $\pm$ 0.024 & 0.243 $\pm$ 0.023 & 0.566 $\pm$ 0.006 & 0.269 $\pm$ 0.005 \\
        KADID-10K~\cite{kadid10k} & \textbf{0.596 $\pm$ 0.005} & \textbf{0.567 $\pm$ 0.011} & \textbf{0.640 $\pm$ 0.001} & \underline{0.448 $\pm$ 0.021} & \underline{0.368 $\pm$ 0.012} & \underline{0.426 $\pm$ 0.009} & 0.598 $\pm$ 0.005 & \underline{0.520 $\pm$ 0.005} \\
        FoMo (\textbf{Ours}) & \underline{0.586 $\pm$ 0.003} & 0.489 $\pm$ 0.003 & \underline{0.512 $\pm$ 0.001} & \textbf{0.526 $\pm$ 0.002} & \textbf{0.547 $\pm$ 0.016} & \textbf{0.473 $\pm$ 0.043} & \underline{0.605 $\pm$ 0.004} & \textbf{0.534 $\pm$ 0.006} \\
    \midrule
    \multicolumn{9}{c}{\textit{Results on CSIQ~\cite{csiq}}} \\
    \midrule
        BAPPS~\cite{lpips} & \underline{0.786 $\pm$ 0.002} & 0.701 $\pm$ 0.003 & 0.654 $\pm$ 0.001 & 0.287 $\pm$ 0.020 & 0.298 $\pm$ 0.011 & 0.315 $\pm$ 0.036 & \textbf{0.741 $\pm$ 0.001} & 0.540 $\pm$ 0.005 \\
        PieAPP~\cite{pieapp} & 0.773 $\pm$ 0.004 & \textbf{0.713 $\pm$ 0.009} & 0.611 $\pm$ 0.004 & 0.229 $\pm$ 0.029 & 0.361 $\pm$ 0.012 & 0.405 $\pm$ 0.009 & \underline{0.724 $\pm$ 0.005} & 0.545 $\pm$ 0.003 \\
        NIGHTS~\cite{dreamsim} & \textbf{0.790 $\pm$ 0.002} & 0.664 $\pm$ 0.001 & -0.384 $\pm$ 0.005 & 0.230 $\pm$ 0.021 & 0.244 $\pm$ 0.016 & 0.291 $\pm$ 0.024 & 0.678 $\pm$ 0.009 & 0.359 $\pm$ 0.004 \\
        KADID-10K~\cite{kadid10k} & 0.768 $\pm$ 0.005 & \underline{0.705 $\pm$ 0.020} & \textbf{0.776 $\pm$ 0.001} & \underline{0.447 $\pm$ 0.007} & \underline{0.430 $\pm$ 0.021} & \underline{0.510 $\pm$ 0.029} & 0.682 $\pm$ 0.012 & \underline{0.617 $\pm$ 0.010} \\
        FoMo (\textbf{Ours}) & 0.781 $\pm$ 0.002 & 0.672 $\pm$ 0.006 & \underline{0.750 $\pm$ 0.001} & \textbf{0.626 $\pm$ 0.002} & \textbf{0.723 $\pm$ 0.019} & \textbf{0.609 $\pm$ 0.030} & 0.709 $\pm$ 0.003 & \textbf{0.696 $\pm$ 0.004} \\
    \midrule
    \multicolumn{9}{c}{\textit{Results on LIVE~\cite{live}}} \\
    \midrule
        BAPPS~\cite{lpips} & \textbf{0.800 $\pm$ 0.001} & \textbf{0.757 $\pm$ 0.005} & 0.644 $\pm$ 0.001 & 0.508 $\pm$ 0.029 & 0.298 $\pm$ 0.043 & 0.203 $\pm$ 0.041 & \underline{0.766 $\pm$ 0.002} & 0.568 $\pm$ 0.008 \\
        PieAPP~\cite{pieapp} & 0.769 $\pm$ 0.018 & 0.746 $\pm$ 0.022 & 0.649 $\pm$ 0.003 & 0.324 $\pm$ 0.018 & 0.288 $\pm$ 0.023 & 0.370 $\pm$ 0.015 & 0.728 $\pm$ 0.005 & 0.554 $\pm$ 0.006 \\
        NIGHTS~\cite{dreamsim} & \underline{0.797 $\pm$ 0.003} & 0.744 $\pm$ 0.004 & -0.574 $\pm$ 0.003 & 0.307 $\pm$ 0.016 & 0.392 $\pm$ 0.045 & 0.256 $\pm$ 0.041 & 0.679 $\pm$ 0.009 & 0.372 $\pm$ 0.012 \\
        KADID-10K~\cite{kadid10k} & 0.789 $\pm$ 0.017 & 0.734 $\pm$ 0.044 & \underline{0.801 $\pm$ 0.001} & \underline{0.580 $\pm$ 0.021} & \underline{0.627 $\pm$ 0.011} & \underline{0.595 $\pm$ 0.015} & 0.713 $\pm$ 0.016 & \underline{0.691 $\pm$ 0.014} \\
        FoMo (\textbf{Ours}) & 0.791 $\pm$ 0.007 & \underline{0.748 $\pm$ 0.007} & \textbf{0.804 $\pm$ 0.001} & \textbf{0.720 $\pm$ 0.003} & \textbf{0.754 $\pm$ 0.017} & \textbf{0.734 $\pm$ 0.038} & \textbf{0.769 $\pm$ 0.004} & \textbf{0.760 $\pm$ 0.005} \\
    \bottomrule
    \end{tabular}
    }
    \label{tab:combined_results_krocc}
\end{table*}


\begin{table*}[htbp]
    \centering
    \caption{Comparison of training datasets and their objectives on four reference-based IQA benchmarks, across CNN- and Transformer-based backbones. Mean $\pm$ standard deviation of PLCC over five random seeds. $^\dagger$Evaluated under 224 resolution.}
    \vspace{5pt}
    \resizebox{\textwidth}{!}{
    \begin{tabular}{l |ccc | cccc | c}
    \toprule
        \multirow{2.5}{*}{\textbf{Train Data}}
        &\multicolumn{3}{c|}{\textbf{CNN-based}} & \multicolumn{4}{c|}{\textbf{Transformer-based}} & \multirow{2.5}{*}{\textbf{Avg.}}\\
        \cmidrule(lr){2-4} \cmidrule(lr){5-8}
        & LPIPS-Alex & LPIPS-VGG & DISTS & DINOv3 & CLIP & MAE & DreamSim$^\dagger$ \\
    \midrule
    \multicolumn{9}{c}{\textit{Results on PIPAL~\cite{pipal}}} \\
    \midrule
        BAPPS~\cite{lpips} & \underline{0.659 $\pm$ 0.005} & \underline{0.676 $\pm$ 0.014} & 0.495 $\pm$ 0.002 & 0.314 $\pm$ 0.015 & 0.186 $\pm$ 0.043 & \underline{0.348 $\pm$ 0.018} & \textbf{0.779 $\pm$ 0.007} & 0.494 $\pm$ 0.010 \\
        PieAPP~\cite{pieapp} & 0.607 $\pm$ 0.010 & 0.637 $\pm$ 0.014 & 0.558 $\pm$ 0.003 & \underline{0.364 $\pm$ 0.014} & 0.284 $\pm$ 0.029 & 0.338 $\pm$ 0.025 & 0.707 $\pm$ 0.005 & \underline{0.499 $\pm$ 0.003} \\
        NIGHTS~\cite{dreamsim} & 0.634 $\pm$ 0.006 & 0.605 $\pm$ 0.007 & 0.361 $\pm$ 0.002 & 0.296 $\pm$ 0.028 & 0.324 $\pm$ 0.030 & 0.259 $\pm$ 0.037 & 0.669 $\pm$ 0.022 & 0.450 $\pm$ 0.008 \\
        KADID-10K~\cite{kadid10k} & 0.599 $\pm$ 0.022 & 0.623 $\pm$ 0.034 & \underline{0.562 $\pm$ 0.001} & 0.351 $\pm$ 0.021 & \underline{0.346 $\pm$ 0.019} & 0.274 $\pm$ 0.034 & 0.619 $\pm$ 0.007 & 0.482 $\pm$ 0.008 \\
        FoMo (\textbf{Ours}) & \textbf{0.768 $\pm$ 0.005} & \textbf{0.732 $\pm$ 0.008} & \textbf{0.656 $\pm$ 0.001} & \textbf{0.706 $\pm$ 0.006} & \textbf{0.652 $\pm$ 0.023} & \textbf{0.625 $\pm$ 0.016} & \underline{0.764 $\pm$ 0.010} & \textbf{0.700 $\pm$ 0.003} \\
    \midrule
    \multicolumn{9}{c}{\textit{Results on TID2013~\cite{tid2013}}} \\
    \midrule
        BAPPS~\cite{lpips} & 0.814 $\pm$ 0.003 & 0.755 $\pm$ 0.002 & 0.664 $\pm$ 0.001 & 0.488 $\pm$ 0.028 & 0.343 $\pm$ 0.037 & 0.330 $\pm$ 0.030 & \textbf{0.850 $\pm$ 0.002} & 0.606 $\pm$ 0.012 \\
        PieAPP~\cite{pieapp} & 0.798 $\pm$ 0.006 & \underline{0.777 $\pm$ 0.005} & 0.743 $\pm$ 0.003 & 0.405 $\pm$ 0.024 & 0.344 $\pm$ 0.020 & 0.513 $\pm$ 0.030 & 0.808 $\pm$ 0.008 & 0.627 $\pm$ 0.006 \\
        NIGHTS~\cite{dreamsim} & 0.805 $\pm$ 0.004 & 0.733 $\pm$ 0.003 & 0.686 $\pm$ 0.003 & 0.391 $\pm$ 0.036 & 0.386 $\pm$ 0.056 & 0.434 $\pm$ 0.045 & 0.791 $\pm$ 0.005 & 0.604 $\pm$ 0.011 \\
        KADID-10K~\cite{kadid10k} & \underline{0.815 $\pm$ 0.007} & \textbf{0.785 $\pm$ 0.014} & \textbf{0.845 $\pm$ 0.001} & \underline{0.694 $\pm$ 0.019} & \underline{0.608 $\pm$ 0.014} & \underline{0.643 $\pm$ 0.027} & 0.825 $\pm$ 0.006 & \underline{0.745 $\pm$ 0.006} \\
        FoMo (\textbf{Ours}) & \textbf{0.817 $\pm$ 0.003} & 0.753 $\pm$ 0.005 & \underline{0.770 $\pm$ 0.001} & \textbf{0.768 $\pm$ 0.002} & \textbf{0.788 $\pm$ 0.018} & \textbf{0.657 $\pm$ 0.046} & \underline{0.831 $\pm$ 0.003} & \textbf{0.769 $\pm$ 0.006} \\
    \midrule
    \multicolumn{9}{c}{\textit{Results on CSIQ~\cite{csiq}}} \\
    \midrule
        BAPPS~\cite{lpips} & 0.945 $\pm$ 0.001 & \textbf{0.908 $\pm$ 0.003} & 0.850 $\pm$ 0.001 & 0.561 $\pm$ 0.036 & 0.483 $\pm$ 0.024 & 0.468 $\pm$ 0.051 & \textbf{0.932 $\pm$ 0.001} & 0.735 $\pm$ 0.007 \\
        PieAPP~\cite{pieapp} & 0.934 $\pm$ 0.004 & \underline{0.907 $\pm$ 0.008} & 0.862 $\pm$ 0.002 & 0.437 $\pm$ 0.044 & 0.594 $\pm$ 0.015 & 0.594 $\pm$ 0.011 & \underline{0.920 $\pm$ 0.004} & 0.750 $\pm$ 0.006 \\
        NIGHTS~\cite{dreamsim} & \textbf{0.945 $\pm$ 0.002} & 0.877 $\pm$ 0.002 & 0.769 $\pm$ 0.003 & 0.450 $\pm$ 0.060 & 0.399 $\pm$ 0.033 & 0.455 $\pm$ 0.042 & 0.878 $\pm$ 0.008 & 0.682 $\pm$ 0.012 \\
        KADID-10K~\cite{kadid10k} & 0.932 $\pm$ 0.004 & 0.889 $\pm$ 0.027 & \textbf{0.934 $\pm$ 0.000} & \underline{0.730 $\pm$ 0.018} & \underline{0.727 $\pm$ 0.029} & \underline{0.735 $\pm$ 0.034} & 0.891 $\pm$ 0.007 & \underline{0.834 $\pm$ 0.011} \\
        FoMo (\textbf{Ours}) & \underline{0.945 $\pm$ 0.001} & 0.888 $\pm$ 0.006 & \underline{0.931 $\pm$ 0.001} & \textbf{0.874 $\pm$ 0.002} & \textbf{0.923 $\pm$ 0.011} & \textbf{0.816 $\pm$ 0.030} & 0.905 $\pm$ 0.003 & \textbf{0.897 $\pm$ 0.004} \\
    \midrule
    \multicolumn{9}{c}{\textit{Results on LIVE~\cite{live}}} \\
    \midrule
        BAPPS~\cite{lpips} & \textbf{0.946 $\pm$ 0.001} & \textbf{0.929 $\pm$ 0.003} & 0.839 $\pm$ 0.000 & 0.760 $\pm$ 0.031 & 0.462 $\pm$ 0.054 & 0.366 $\pm$ 0.035 & \underline{0.935 $\pm$ 0.001} & 0.748 $\pm$ 0.011 \\
        PieAPP~\cite{pieapp} & 0.923 $\pm$ 0.016 & 0.916 $\pm$ 0.016 & 0.870 $\pm$ 0.002 & 0.493 $\pm$ 0.023 & 0.469 $\pm$ 0.047 & 0.550 $\pm$ 0.021 & 0.912 $\pm$ 0.002 & 0.733 $\pm$ 0.011 \\
        NIGHTS~\cite{dreamsim} & \underline{0.946 $\pm$ 0.002} & 0.922 $\pm$ 0.003 & 0.753 $\pm$ 0.004 & 0.564 $\pm$ 0.020 & 0.586 $\pm$ 0.062 & 0.402 $\pm$ 0.055 & 0.883 $\pm$ 0.006 & 0.722 $\pm$ 0.015 \\
        KADID-10K~\cite{kadid10k} & 0.935 $\pm$ 0.018 & 0.903 $\pm$ 0.036 & \underline{0.947 $\pm$ 0.000} & \underline{0.816 $\pm$ 0.016} & \underline{0.853 $\pm$ 0.008} & \underline{0.809 $\pm$ 0.013} & 0.906 $\pm$ 0.010 & \underline{0.881 $\pm$ 0.011} \\
        FoMo (\textbf{Ours}) & 0.941 $\pm$ 0.004 & \underline{0.923 $\pm$ 0.004} & \textbf{0.950 $\pm$ 0.001} & \textbf{0.913 $\pm$ 0.001} & \textbf{0.928 $\pm$ 0.011} & \textbf{0.913 $\pm$ 0.027} & \textbf{0.935 $\pm$ 0.002} & \textbf{0.929 $\pm$ 0.004} \\
    \bottomrule
    \end{tabular}
    }
    \label{tab:combined_results_plcc}
\end{table*}

\begin{table*}[htbp]
    \centering
    \caption{Comparative experiment on the objective function and label type. Ranked binary cross-entropy loss (Rank) consistently shows stronger results, compared to triplet-based paired cross-entropy loss (2AFC) and L1 regression loss on the label. Our label has shown to be more helpful in training, compared to the large-scale human annotated KADID-10k~\cite{kadid10k}. PIPAL SROCC, mean $\pm$ standard deviation over five random seeds.}
    \vspace{5pt}
    \resizebox{\textwidth}{!}{
    \begin{tabular}{l | c c c | c c c | c c c}
    \toprule
        & \multicolumn{6}{c|}{\textbf{FoMo (Ours)}} & \multicolumn{3}{c}{\textbf{KADID-10K~\cite{kadid10k}}} \\
        \cmidrule(lr){2-7} \cmidrule(lr){8-10}
        \textbf{Label} & \multicolumn{3}{c|}{$t$} & \multicolumn{3}{c|}{$s$} & \multicolumn{3}{c}{DMOS} \\
        \cmidrule(lr){2-4} \cmidrule(lr){5-7} \cmidrule(lr){8-10}
        \textbf{Objective} & Rank & 2AFC & L1 & Rank & 2AFC & L1 & Rank & 2AFC & L1 \\
    \midrule
        LPIPS-Alex~\cite{lpips} & 0.733 $\pm$ 0.006 & 0.592 $\pm$ 0.002 & 0.440 $\pm$ 0.059 & 0.733 $\pm$ 0.006 & 0.592 $\pm$ 0.002 & 0.390 $\pm$ 0.049 & 0.625 $\pm$ 0.004 & 0.611 $\pm$ 0.003 & 0.577 $\pm$ 0.025 \\
        DINOv3~\cite{dinov3} & 0.699 $\pm$ 0.006 & 0.622 $\pm$ 0.020 & 0.694 $\pm$ 0.008 & 0.697 $\pm$ 0.006 & 0.614 $\pm$ 0.009 & 0.695 $\pm$ 0.013 & 0.309 $\pm$ 0.030 & 0.145 $\pm$ 0.010 & 0.298 $\pm$ 0.010 \\
    \bottomrule
    \end{tabular}
    }
    \label{tab:ablation_pred_type_full}
\end{table*}

\input{02_related_work}

\section{Human Alignment Experiment}
\begin{figure}
    \centering
    \includegraphics[width=\linewidth]{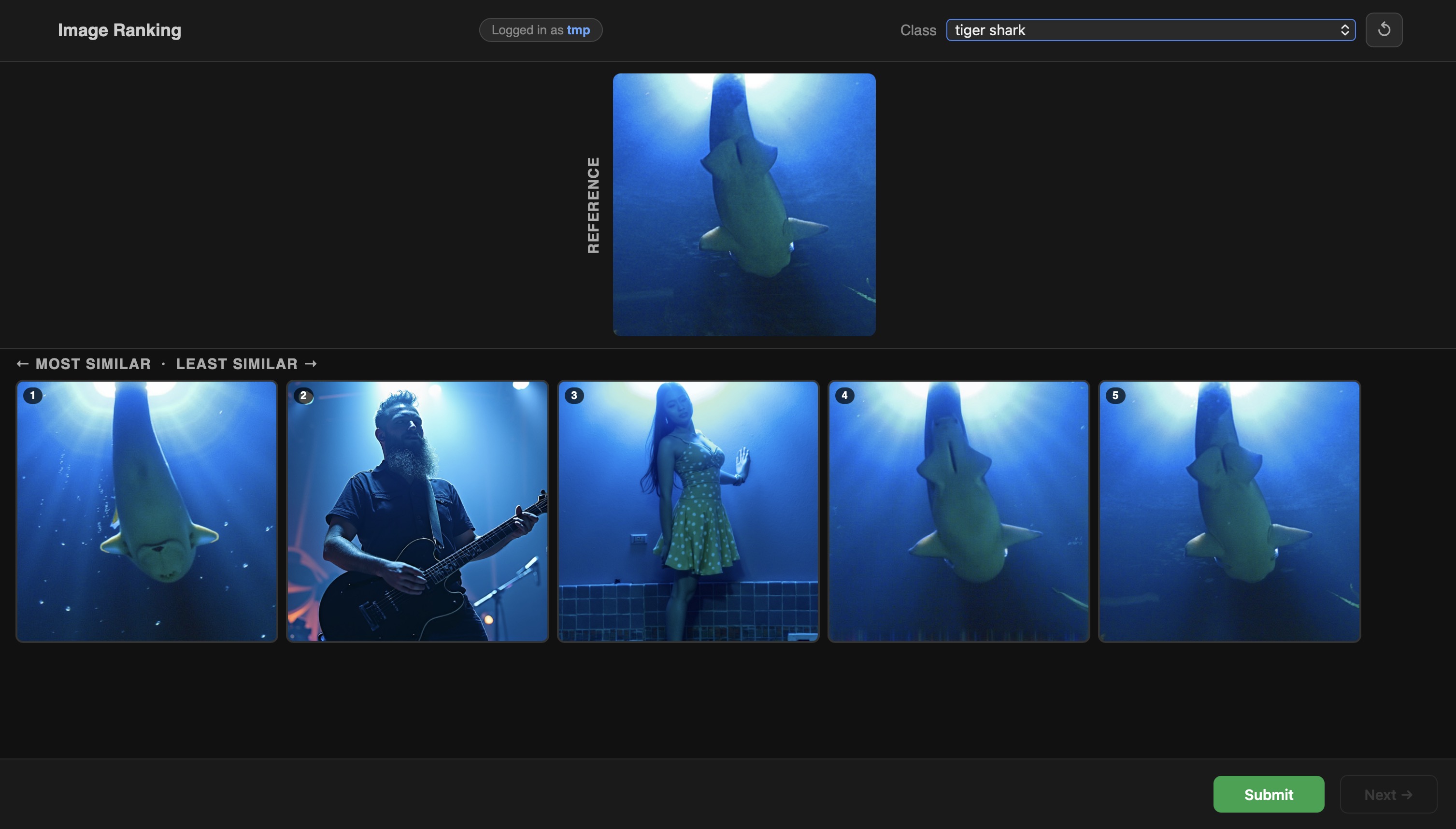}
    \caption{Screenshot of the human annotation website for experiments in Sec.~\ref{sec:emp_grounding}}
    \label{fig:annotation_ui1}
\end{figure}

\begin{figure}
    \centering
    \includegraphics[width=\linewidth]{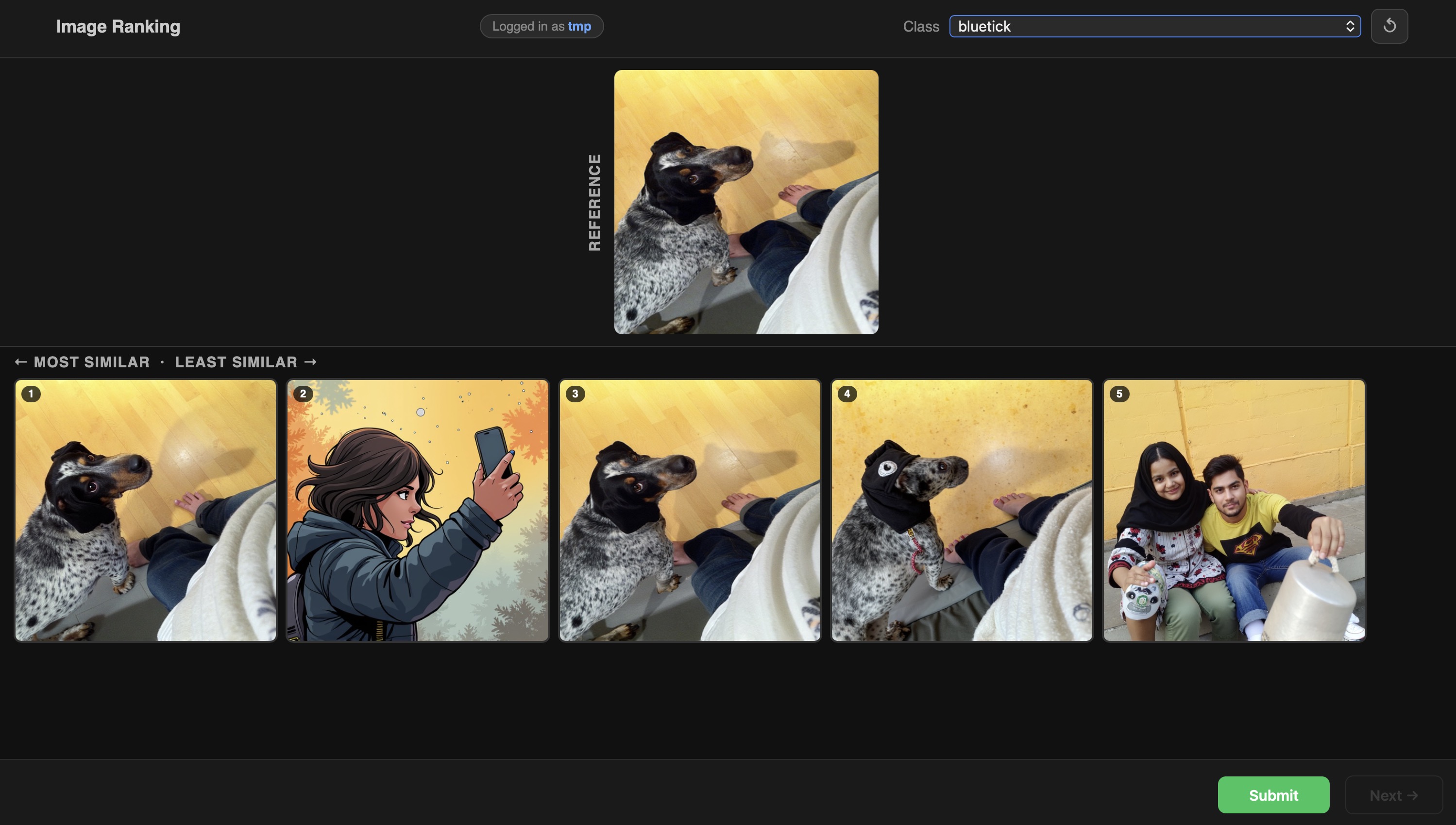}
    \caption{Screenshot of the human annotation website for experiments in Sec.~\ref{sec:emp_grounding}}
    \label{fig:annotation_ui2}
\end{figure}

\begin{figure}
    \centering
    \includegraphics[width=0.77\linewidth]{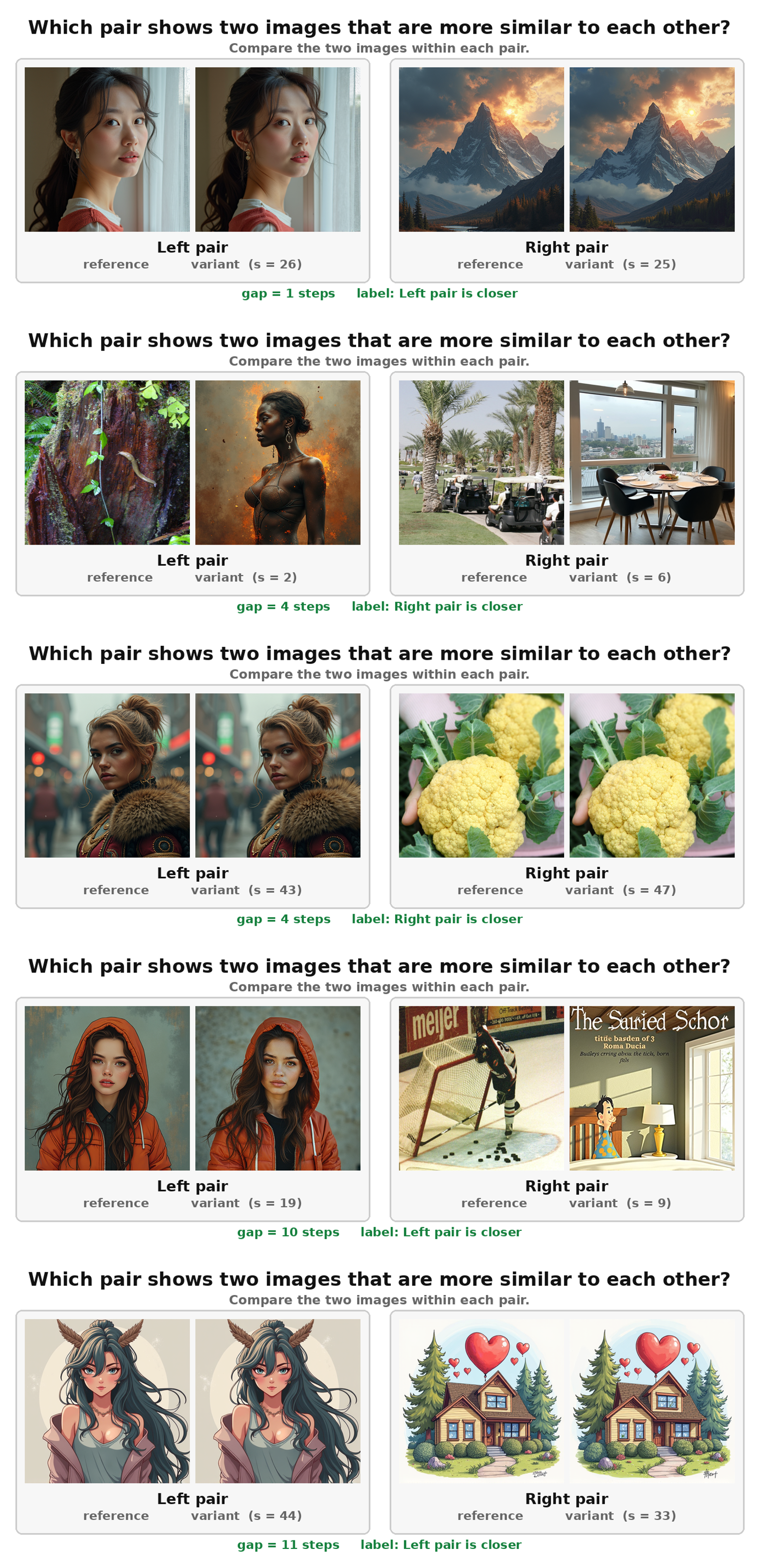}
    \caption{Examples from cross-reference human study, with each samples' forked moments, the gap of forking moments and labels from FoMo provided.}
    \label{fig:cross_ref_example_full_part1}
\end{figure}

\begin{figure}
    \centering
    \includegraphics[width=0.77\linewidth]{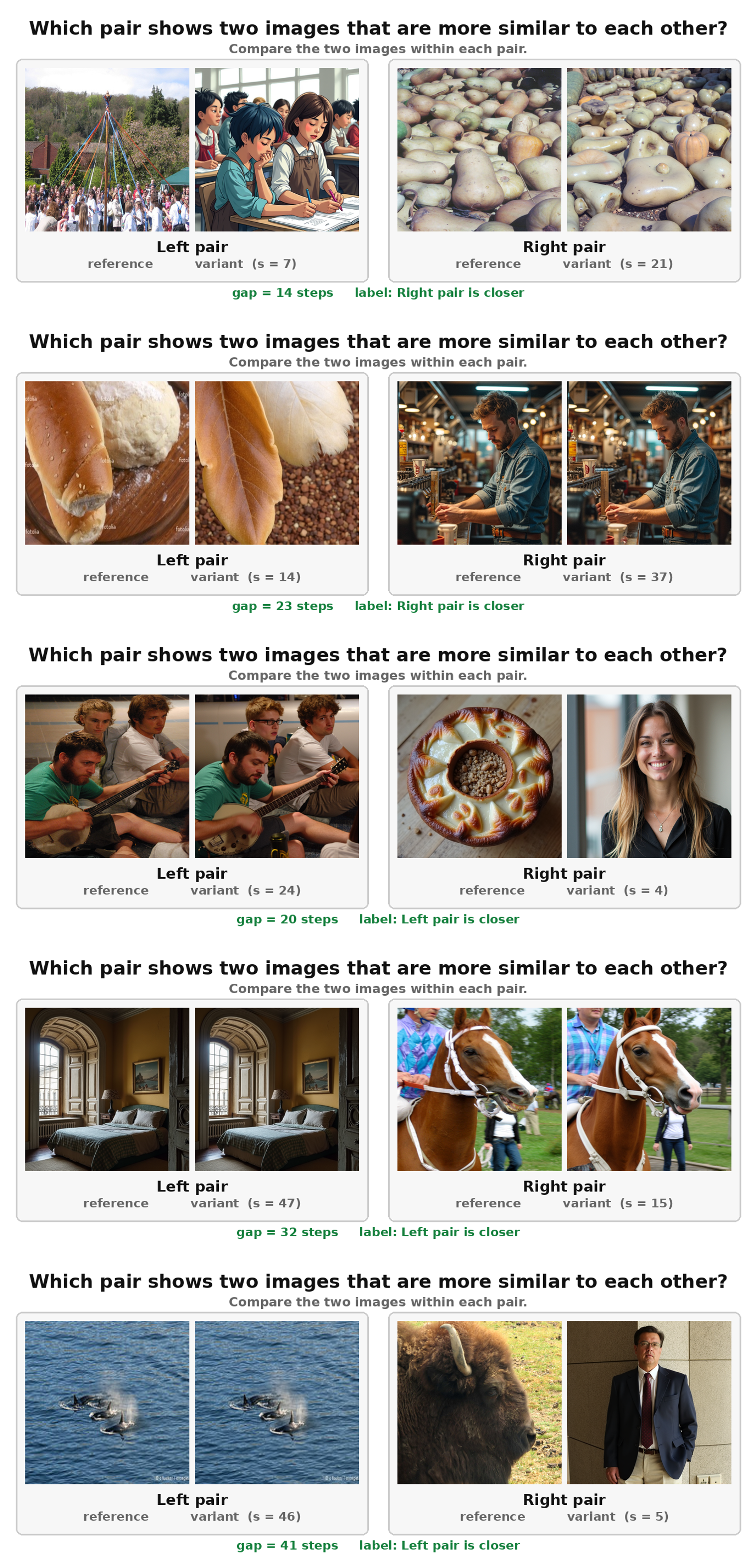}
    \caption{Examples from cross-reference human study, with each samples' forked moments, the gap of forking moments and labels from FoMo provided.}
    \label{fig:cross_ref_example_full_part2}
\end{figure}

We present the web user-interface used for single-reference human study from the experiment of Sec.~\ref{sec:single_reference_study} are in Fig.~\ref{fig:annotation_ui1} and ~\ref{fig:annotation_ui2}, and additional examples of experiment from Sec.~\ref{sec:cross_reference_study} are in Fig.~\ref{fig:cross_ref_example_full_part1} and ~\ref{fig:cross_ref_example_full_part2}.


\section{Data Augmentation}
One of a key advantage of our approach is its compatibility with a substantially broader range of data augmentation strategies than human-annotated alternatives permit. Pairwise preference datasets such as BAPPS~\cite{lpips} and NIGHTS~\cite{dreamsim} impose strict constraints on applicable augmentations: while label-preserving transformations such as random horizontal flips and in-plane rotations can be safely applied without altering perceptual judgments, aggressive spatial augmentations, most notably random resize cropping, are not applicable. Annotations in these datasets reflect global perceptual preferences elicited at a fixed resolution over entire image triplets; a crop that exposes only a local region may induce a different perceptual ordering and thereby contradict the original annotation.

Our approach enjoys better flexibility because the forking moment label is derived from a pre-defined noise schedule and can therefore be recomputed analytically under any spatial transformation. We describe how two canonical augmentations are accommodated within our framework. Both derivations rely on the established result that the diffusion noise schedule must be rescaled with image resolution~\cite{simplediffusion, importanceofnoiseschedule}. Concretely, the log signal-to-noise ratio (log-SNR) of the variance-preserving forward process shifts with resolution as
\begin{equation}
    \log\text{SNR}(t;r) = \log\text{SNR}(t;r_0) + 2 \log(\frac{r_0}{r}),
    \label{eq:sch_shift}
\end{equation}
where $r$ denotes the image resolution (\textit{e.g.}, the shorter spatial dimension in pixels), $r_0$ is a reference resolution at which the baseline schedule is defined, and $t \in [0, 1]$ is the continuous interpolation factor. Given the variance-preserving constraint $a_t^2 + b_t^2 = 1$, the forward-process noise coefficients at resolution $r$ are recovered as
\begin{equation}
    a^2_t(r) = \sigma(\log \text{SNR}(t; r)),
    b^2_t(r) = 1 - a^2_t(r),
\end{equation}
where $\sigma(\cdot)$ denote the sigmoid operator..
We refer to the log-SNR value at the forking moment as $\lambda_s = \log\text{SNR}(t_0;\,r_0)$, which serves as a resolution-invariant proxy for perceptual divergence in both cases below.

\paragraph{Resizing} Rescaling an image from $r_0$ to a new resolution $r_\text{resize}$ preserves the global scene content and structure. The relative noise level at which content is destroyed therefore remains unchanged, and so does the perceptual divergence between the reference and distorted images.
Consequently, the forking timestep $s$ is retained as the label after resizing. However, because the noise schedule is resolution-dependent (Eq.~\ref{eq:sch_shift}), the continuous interpolation factor $t$ corresponding to $s$ shifts implicitly. If $s$ maps to interpolation factor $t_0$ under the original schedule, then at resolution $r_\text{resize}$ the factor $t_0$ must be updated to $t_\text{resize}$ such that
\begin{equation}
    \log\text{SNR}(t_\text{resize}; r_\text{resize}) = \log\text{SNR}(t_0, r_0),
\end{equation}
which amounts to a shift of $t_0$ by $2\log(r_0/r_\text{resize})$ in log-SNR space.

\paragraph{Random Resize Cropping} Cropping simultaneously alters the spatial resolution and the visible image content. Since the cropped region depicts only a portion of the original scene, the forking timestep $s$ can no longer be assumed invariant.
However, the continuous interpolation factor $t$, which encodes the relative signal-to-noise level at which the two images perceptually diverge, is a resolution-agnostic quantity and is preserved across the crop. This follows directly from the pixel-wise nature of the diffusion forward process: a noised image at step $s$ is formed as $x_{s} = a_{t_0}x_0 + b_{t_0}\epsilon$, where the mixing ratio $t_0$ is applied uniformly across all spatial locations. Any crop of $x_{s}$ is therefore a crop of the same mixture at factor $t_0$, irrespective of the global image extent or resolution.

The label for the cropped region at resolution $r_\text{crop}$ is recomputed as follows. The interpolation factor $t_0$ is retained from the original annotation, and $\lambda_s$ is computed as above. The noise schedule is then rescaled to resolution $r_\text{crop}$ via Eq.~\ref{eq:sch_shift}, yielding a new log-SNR curve. Finally, the forking timestep $s_\text{crop}$ is recovered by identifying the discrete step $s \in \{0, \ldots, S{-}1\}$ whose normalized time $s/S$ maps to log-SNR closest to $\lambda_s$ under the rescaled schedule. In the continuous-time limit this reduces to an exact inversion of the log-SNR function at $r_\text{crop}$.

\label{sec:random_resize_crop}
In summary, resizing preserves the forking timestep $s$ while the interpolation factor $t$ shifts, whereas random resize cropping preserves $t$ while $s$ must be recomputed. In both cases the true invariant is $\lambda_s$: the log-SNR at the moment of perceptual divergence.

\section{Comparison against Off-the-Shelf Metrics}
\label{app:offshelf}
Directly comparing off-the-shelf metrics to ours could make it hard to strictly ablate the benefits of our proposed approach.
Thus, the results in the main paper, such as Table~\ref{tab:combined_results} is intended to be ablative. Most of the experimental settings are shared, from frozen backbone, prediction head initialization, optimizer, schedule and budget, etc., leaving the training data and the objective as the only variables.
Yet, how our approach performs in comparison to off-the-shelf metrics is nonetheless worth establishing, and Table~\ref{tab:offshelf} presents the results, placing the public LPIPS-Alex, LPIPS-VGG, DISTS and DreamSim
checkpoints beside the same architectures trained with FoMo supervision and
evaluated under the same protocol. (In DreamSim, all images were resized to 224 resolution, following the protocol of DreamSim.)

Across the four architectures, FoMo supervision is comparable to the released checkpoints and slightly ahead overall, leading in half of the cells (25 of 48) and in all twelve for LPIPS-Alex.
It is worth noting that this result is obtained under a single configuration applied to every backbone, shared unchanged across the three CNN backbones, with nothing tuned per architecture or per benchmark and without any human-annotated supervision.
The released checkpoints, by contrast, each reflect considerable per-metric care: their own design choices, hyper-parameters and training sets.
This result again proves the strength of FoMo, and further implies the potential that the metrics could still improve more with careful tuning of hyper-parameters for each model.

\begin{table}[htbp]
    \centering
    \caption{Released off-the-shelf metrics compared with the same architecture trained with FoMo supervision. Better of each pair in \textbf{bold}. Released checkpoints are single deterministic models whereas FoMo columns are 5-seed means.}
    \vspace{5pt}
    \resizebox{\textwidth}{!}{
    \begin{tabular}{ll | cc | cc | cc | cc}
    \toprule
        & & \multicolumn{2}{c|}{\textbf{LPIPS-Alex}~\cite{lpips}} & \multicolumn{2}{c|}{\textbf{LPIPS-VGG}~\cite{lpips}} & \multicolumn{2}{c|}{\textbf{DISTS}~\cite{dists}} & \multicolumn{2}{c}{\textbf{DreamSim}~\cite{dreamsim}} \\
        \cmidrule(lr){3-4} \cmidrule(lr){5-6} \cmidrule(lr){7-8} \cmidrule(lr){9-10}
        & & Released & FoMo & Released & FoMo & Released & FoMo & Released & FoMo \\
    \midrule
    \multirow{3}{*}{PIPAL~\cite{pipal}}
        & SROCC & 0.620 & \textbf{0.733} & 0.612 & \textbf{0.683} & \textbf{0.672} & 0.615 & 0.759 & \textbf{0.776} \\
        & KROCC & 0.435 & \textbf{0.536} & 0.438 & \textbf{0.497} & \textbf{0.482} & 0.435 & 0.563 & \textbf{0.571} \\
        & PLCC  & 0.623 & \textbf{0.767} & 0.668 & \textbf{0.712} & \textbf{0.685} & 0.654 & \textbf{0.780} & 0.762 \\
    \midrule
    \multirow{3}{*}{TID2013~\cite{tid2013}}
        & SROCC & 0.745 & \textbf{0.785} & \textbf{0.670} & 0.663 & \textbf{0.708} & 0.691 & \textbf{0.812} & 0.801 \\
        & KROCC & 0.548 & \textbf{0.586} & \textbf{0.497} & 0.489 & \textbf{0.521} & 0.512 & \textbf{0.614} & 0.605 \\
        & PLCC  & 0.753 & \textbf{0.816} & \textbf{0.749} & 0.746 & 0.755 & \textbf{0.766} & 0.746 & \textbf{0.831} \\
    \midrule
    \multirow{3}{*}{CSIQ~\cite{csiq}}
        & SROCC & 0.923 & \textbf{0.938} & \textbf{0.883} & 0.859 & \textbf{0.930} & 0.918 & \textbf{0.911} & 0.894 \\
        & KROCC & 0.750 & \textbf{0.781} & \textbf{0.697} & 0.672 & \textbf{0.764} & 0.750 & \textbf{0.738} & 0.709 \\
        & PLCC  & 0.920 & \textbf{0.944} & \textbf{0.906} & 0.888 & \textbf{0.938} & 0.931 & \textbf{0.928} & 0.900 \\
    \midrule
    \multirow{3}{*}{LIVE~\cite{live}}
        & SROCC & 0.924 & \textbf{0.948} & \textbf{0.932} & 0.923 & 0.948 & \textbf{0.954} & 0.910 & \textbf{0.931} \\
        & KROCC & 0.751 & \textbf{0.791} & \textbf{0.765} & 0.748 & 0.793 & \textbf{0.804} & 0.745 & \textbf{0.769} \\
        & PLCC  & 0.916 & \textbf{0.940} & \textbf{0.934} & 0.923 & 0.945 & \textbf{0.949} & 0.918 & \textbf{0.934} \\
    \bottomrule
    \end{tabular}
    }
    \label{tab:offshelf}
\end{table}

\section{Per-Sample Label Variance}
\label{app:variance}
The forking construction is stochastic: two variants generated from the same
reference at the same forking step are not identical, because the noise
re-injected at the fork differs, yet both carry the same label. To quantify the
resulting spread we take 120 ImageNet references, generate $K = 8$ variants at
each of five forking steps, 4,800 images in total, and measure every
variant's distance to its reference. For one reference at one forking step this
gives eight distances, of which we take the mean and the standard deviation.
Table~\ref{tab:variance} reports both averaged over the 120 references, together
with their ratio, the coefficient of variation (CoV). The spread is small in every
regime: the standard deviation stays below $0.05$ in absolute terms and at most
$10.1\%$ of the distance it accompanies, while the mean distance itself
changes six- to eightfold across the schedule. Re-running the generator
therefore perturbs a sample by far less than the label separates it from its
neighbors, and the perturbation reorders two variants of the same reference in
at most $2.8\%$ (LPIPS-Alex) and $7.0\%$ (DISTS) of comparisons. Since this
noise is independent across the 480k training pairs and averages out over them,
we regard it as negligible for training.

\begin{table}[htbp]
    \centering
    \caption{Spread of the measured distance across generation seeds. 120
    references $\times$ 5 forking steps $\times$ $K = 8$ seeds. For each
    reference we take the eight distances obtained at one forking step and
    compute their mean and standard deviation; the table reports these averaged
    over the 120 references, with CoV their ratio. A larger $s$ denotes a later
    fork, and hence a variant closer to the reference.}
    \vspace{5pt}
    \begin{tabular}{c | c c c | c c c}
    \toprule
        \multirow{2.5}{*}{\textbf{Forking step} $s$} & \multicolumn{3}{c|}{\textbf{LPIPS-Alex}} & \multicolumn{3}{c}{\textbf{DISTS}} \\
        \cmidrule(lr){2-4} \cmidrule(lr){5-7}
        & mean $d$ & std & CoV & mean $d$ & std & CoV \\
    \midrule
        5  & 0.692 & 0.043 & 6.4\% & 0.391 & 0.034 & 8.7\%  \\
        15 & 0.518 & 0.046 & 9.1\% & 0.306 & 0.031 & 10.1\% \\
        25 & 0.371 & 0.030 & 7.9\% & 0.225 & 0.021 & 9.2\%  \\
        35 & 0.233 & 0.011 & 4.7\% & 0.150 & 0.011 & 6.9\%  \\
        45 & 0.090 & 0.002 & 2.7\% & 0.067 & 0.004 & 5.6\%  \\
    \bottomrule
    \end{tabular}
    \label{tab:variance}
\end{table}

\section{Per-Distortion-Type Analysis}
\label{app:pertype}
Table~\ref{tab:combined_results} reports one SROCC per benchmark. This appendix asks which distortion families that number is built from.
We recompute SROCC within each distortion type of all four benchmarks and compare FoMo against the strongest human-annotated recipe for the same backbone on the same benchmark.
PIPAL is evaluated on its training split, the only one whose distortion types are identifiable, so its full-set values are not comparable to the validation numbers of Table~\ref{tab:combined_results}.

With the Transformer backbone, FoMo supervision leads on almost every distortion family of every benchmark (Table~\ref{tab:pertype}). The gain is largest exactly where the backbone on its own is weakest, the super-resolution families of PIPAL, and the noise and contrast families of CSIQ. So the supervision, not the architecture, is what supplies the perceptual ordering. 
With the CNN backbone the picture is narrower, as one would expect of a network whose ImageNet features already encode a perceptual prior.
FoMo leads throughout PIPAL, but on the three legacy synthetic-distortion benchmarks it trails on a majority of families, by margins of hundredths of a point.

The failures are consistent across both backbones and concentrate in three families (Table~\ref{tab:pertype_full} lists every type): additive pixel noise, corruption confined to a small region, and global photometric shifts such as contrast change and mean shift. None of these occurs in our training data.
A diffusion model re-synthesizes an image as a whole, so it never adds pixel-independent noise, never corrupts an isolated rectangle, and never applies a purely photometric change; supervision cannot teach what it never shows.

\begin{table}[htbp]
    \centering
    \caption{Per-distortion-type summary on all four benchmarks. ``Best
    baseline'' is the strongest of the four human-annotated recipes for that
    backbone on that benchmark, selected independently per benchmark. ``Full''
    is the score over the whole benchmark and ``led'' the number of distortion
    types on which FoMo is ahead. PIPAL uses its training split, the only one
    that exposes distortion types. Symmetrized native-resolution protocol,
    seed-0 checkpoints.}
    \vspace{5pt}
    \resizebox{\columnwidth}{!}{
    \begin{tabular}{l | c | c c c | c c c}
    \toprule
        \multirow{2.5}{*}{\textbf{Benchmark}} & \multirow{2.5}{*}{\textbf{types}} & \multicolumn{3}{c|}{\textbf{LPIPS-Alex}} & \multicolumn{3}{c}{\textbf{DINOv3}} \\
        \cmidrule(lr){3-5} \cmidrule(lr){6-8}
        & & best baseline & FoMo & led & best baseline & FoMo & led \\
    \midrule
        PIPAL & 7 & 0.611 & \textbf{0.681} & 7/7 & 0.259 & \textbf{0.525} & 7/7 \\
        TID2013 & 24 & \textbf{0.794} & 0.783 & 10/24 & 0.592 & \textbf{0.710} & 22/24 \\
        CSIQ & 6 & \textbf{0.945} & 0.936 & 1/6 & 0.613 & \textbf{0.808} & 6/6 \\
        LIVE & 5 & \textbf{0.952} & 0.942 & 2/5 & 0.748 & \textbf{0.896} & 5/5 \\
    \bottomrule
    \end{tabular}
    }
    \label{tab:pertype}
\end{table}

\begin{table}[htbp]
    \centering
    \caption{Every distortion type of all four benchmarks, against the strongest
    human-annotated recipe for the same backbone on the same benchmark. Bold
    marks the better of each pair. FoMo supervision leads on 20 of the 42 types
    with the CNN backbone and on 40 of the 42 with the Transformer backbone.}
    \vspace{5pt}
    \small
    \begin{tabular}{l | c c | c c}
    \toprule
        \multirow{2.5}{*}{\textbf{Distortion type}} & \multicolumn{2}{c|}{\textbf{LPIPS-Alex}} & \multicolumn{2}{c}{\textbf{DINOv3}} \\
        \cmidrule(lr){2-3} \cmidrule(lr){4-5}
        & best baseline & FoMo (\textbf{Ours}) & best baseline & FoMo (\textbf{Ours}) \\
    \midrule
    \multicolumn{5}{c}{\textit{PIPAL~\cite{pipal} (training split)}} \\
    \midrule
        SR (traditional) & 0.577 & \textbf{0.669} & 0.098 & \textbf{0.618} \\
        SR (PSNR-oriented) & 0.707 & \textbf{0.785} & 0.315 & \textbf{0.721} \\
        SR (kernel mismatch) & 0.560 & \textbf{0.650} & 0.349 & \textbf{0.537} \\
        SR (GAN-based) & 0.501 & \textbf{0.565} & 0.172 & \textbf{0.471} \\
        Denoising & 0.688 & \textbf{0.757} & 0.437 & \textbf{0.693} \\
        Mixture & 0.570 & \textbf{0.665} & 0.370 & \textbf{0.599} \\
        Traditional & 0.586 & \textbf{0.628} & 0.125 & \textbf{0.346} \\
    \midrule
    \multicolumn{5}{c}{\textit{TID2013~\cite{tid2013}}} \\
    \midrule
        Additive Gaussian noise & \textbf{0.809} & 0.766 & 0.432 & \textbf{0.808} \\
        Additive noise, colour comp. & \textbf{0.742} & 0.690 & 0.352 & \textbf{0.735} \\
        Spatially correlated noise & 0.717 & \textbf{0.744} & 0.659 & \textbf{0.794} \\
        Masked noise & \textbf{0.785} & 0.770 & 0.133 & \textbf{0.608} \\
        High-frequency noise & \textbf{0.847} & 0.806 & 0.506 & \textbf{0.848} \\
        Impulse noise & \textbf{0.552} & 0.527 & 0.589 & \textbf{0.633} \\
        Quantisation noise & \textbf{0.786} & 0.764 & 0.657 & \textbf{0.828} \\
        Gaussian blur & 0.929 & \textbf{0.931} & 0.488 & \textbf{0.785} \\
        Image denoising & 0.857 & \textbf{0.871} & 0.740 & \textbf{0.868} \\
        JPEG & \textbf{0.897} & 0.887 & 0.710 & \textbf{0.889} \\
        JPEG2000 & 0.914 & \textbf{0.934} & 0.756 & \textbf{0.881} \\
        JPEG transmission errors & 0.882 & \textbf{0.898} & 0.645 & \textbf{0.803} \\
        JPEG2000 transmission errors & \textbf{0.799} & 0.791 & 0.568 & \textbf{0.696} \\
        Non-eccentricity pattern noise & 0.782 & \textbf{0.822} & 0.279 & \textbf{0.800} \\
        Local block-wise distortion & 0.335 & \textbf{0.349} & \textbf{0.401} & 0.277 \\
        Mean shift & \textbf{0.778} & 0.737 & 0.126 & \textbf{0.576} \\
        Contrast change & \textbf{0.434} & 0.410 & \textbf{-0.029} & -0.159 \\
        Colour saturation change & \textbf{0.791} & 0.782 & 0.271 & \textbf{0.757} \\
        Multiplicative Gaussian noise & \textbf{0.742} & 0.691 & 0.478 & \textbf{0.748} \\
        Comfort noise & 0.874 & \textbf{0.877} & 0.628 & \textbf{0.890} \\
        Lossy compression of noisy img. & \textbf{0.914} & 0.901 & 0.734 & \textbf{0.885} \\
        Colour quantisation with dither & \textbf{0.812} & 0.786 & 0.592 & \textbf{0.837} \\
        Chromatic aberrations & 0.880 & \textbf{0.890} & 0.605 & \textbf{0.783} \\
        Sparse sampling and reconstr. & 0.925 & \textbf{0.943} & 0.828 & \textbf{0.915} \\
    \midrule
    \multicolumn{5}{c}{\textit{CSIQ~\cite{csiq}}} \\
    \midrule
        AWGN & \textbf{0.940} & 0.913 & 0.459 & \textbf{0.891} \\
        Gaussian blur & \textbf{0.960} & 0.956 & 0.625 & \textbf{0.941} \\
        Contrast change & \textbf{0.949} & 0.929 & 0.076 & \textbf{0.863} \\
        Pink noise & \textbf{0.945} & 0.917 & 0.593 & \textbf{0.891} \\
        JPEG & \textbf{0.958} & 0.949 & 0.800 & \textbf{0.955} \\
        JPEG2000 & 0.942 & \textbf{0.945} & 0.781 & \textbf{0.947} \\
    \midrule
    \multicolumn{5}{c}{\textit{LIVE~\cite{live}}} \\
    \midrule
        Fast fading & 0.962 & \textbf{0.965} & 0.704 & \textbf{0.960} \\
        Gaussian blur & 0.962 & \textbf{0.964} & 0.489 & \textbf{0.903} \\
        JPEG2000 & \textbf{0.950} & 0.941 & 0.711 & \textbf{0.933} \\
        JPEG & \textbf{0.965} & 0.959 & 0.841 & \textbf{0.962} \\
        White noise & \textbf{0.961} & 0.876 & 0.908 & \textbf{0.966} \\
    \bottomrule
    \end{tabular}
    \label{tab:pertype_full}
\end{table}

\section{Limitations}
\label{app:limitations}
Despite the excellent performance in existing benchmarks. 
Our approach has a few limitations.
One major limitation is that, since our approach generated distorted images with diffusion models, the metric models trained with our data and objective may fail in unseen image domains that are out-of-distribution, such as artificial distortions uncommon in nature. For scenarios like those, one could train a diffusion model on the new domain, and use it the trained diffusion model for generating new samples to train in that domain. This way, our approach can overcome its own limitation.

\section{Ethics Statement}
The human annotation study involved only perceptual preference judgments on image pairs, with no deception, no collection of personally identifiable information, and no risk beyond normal screen use. Participants were informed of the task nature prior to annotation. The study was determined exempt from formal IRB review under the minimal-risk behavioral research exemption.

\section{Broader Impacts}
\paragraph{Positive Societal Impact}
Automating perceptual label generation reduces reliance on costly human annotation, lowering the barrier to building human-aligned IQA metrics. Better metrics improve evaluation pipelines across image restoration and synthesis, benefiting applications in medical imaging, compression, and accessibility.

\paragraph{Negative Societal Impact}
More accurate perceptual metrics could be exploited to optimize generative models toward visually convincing outputs that conceal manipulations, potentially aiding synthetic media misuse. The metric may also inherit perceptual biases from the diffusion model used to generate training labels.

%% file: 02_related_work.tex
\section{Related Work}
\subsection{Reference-based Image Quality Assessment}
A reliable perceptual metric must produce a globally consistent quality ordering, one that mirrors how humans rank distortions across the full image set, not just on isolated pairs. This requirement has driven the evolution of IQA annotation methodology.
Early benchmark datasets such as LIVE~\cite{live}, CSIQ~\cite{csiq}, TID2013~\cite{tid2013} and KADID-10k~\cite{kadid10k} collected Mean Opinion Scores (MOS) as ground truth, implicitly assuming that averaged absolute ratings constitute a reliable global quality scale. PieAPP~\cite{pieapp} challenged this directly, arguing that MOS-based annotations are unreliable because absolute quality ratings are inconsistent across raters and sessions: observers apply different scale calibrations and anchor their ratings differently, making cross-image comparisons ambiguous. This motivated a shift toward 2-alternative-forced-choice (2AFC) annotation, where raters compare two distortions relative to a reference rather than score in isolation. Concurrent to this finding, LPIPS~\cite{lpips} also adopts the 2AFC paradigm for labeling and constructs a large-scale dataset gathered from human annotations. PIPAL~\cite{pipal} extends it to GAN-based restorations via Elo-ranked pairwise judgments to expand the coverage of IQA evaluations. DreamSim~\cite{dreamsim} also collects a new dataset based on the 2AFC paradigm, but its main focus is on mid-level similarity rather than low-level similarity.

Despite this recent popularity of 2AFC-style labeling and training, the 2AFC setting carries its own structural weakness. Talebi et al.~\cite{ranksmoothed} point out that mini-batch pairwise optimization never explicitly sees the global ranking of images, and each gradient step accounts for only a small fraction of all possible comparisons; they demonstrate that regularizing with rank-centrality aggregation consistently improves human preference prediction. dipIQ~\cite{dipiq} reinforces this directly by comparing pairwise RankNet and listwise ListNet training objectives on the same automatically generated quality-discriminable pairs, finding that the listwise variant consistently outperforms its pairwise counterpart, a clear empirical signal that optimizing global ordinal structure is superior to aggregating independent pairwise decisions. Yet all of these methods remain bottlenecked by human annotation cost and coverage.

\subsection{Diffusion Models as Perceptual Signal}
There have been several efforts in exploiting diffusion models for perceptual similarity.
DIFT~\cite{dift} and Diffusion Hyperfeatures~\cite{diffhyperfeats} employ diffusion features for structural similarity, but their methods are targeted for geometric correspondence task, rather than low-level perceptual distance.
DiffSim~\cite{diffsim} uses attention layer features in Stable Diffusion~\cite{ldm} to measure visual similarity, but targets style and instance-level consistency in generative customization settings rather than low-level perceptual fidelity.
AnoDDPM~\cite{anoddpm} partially diffuses an image to an intermediate timestep and measures the pixel-level reconstruction divergence after denoising as an anomaly score, but restricted to detecting distributional outliers in medical images.
DifFIQA~\cite{diffiqa} applies perturbation-robustness under DDPM noising as a face image quality signal, scoring faces by the shift in identity embedding between input and reconstructed image, an NR-IQA approach specific to facial content.

Our approach differs from all of the above on two axes. First, unlike feature-based approaches, our approach does not use diffusion model representations at inference time; the diffusion model is used only for generating the data samples and their labels. Second, unlike scoring approaches like AnoDDPM and DifFIQA, our approach uses the denosing generative process as a label generation mechanism, instead of using the generation result in distance computation.